%% file: main.tex
\documentclass{article} % For LaTeX2e
\usepackage{iclr2026_conference,times}

\input{math_commands.tex}

\usepackage{url}
\usepackage{graphicx}
\usepackage{booktabs}
\usepackage{multirow}
\usepackage{caption}
\usepackage[justification=justified,singlelinecheck=false]{caption}
\usepackage{makecell} 
\usepackage{hyperref}
\usepackage{float}
\usepackage{wrapfig}

\newlength\savewidth
\newcommand{\tablestyle}[2]{\setlength{\tabcolsep}{#1}\renewcommand{\arraystretch}{#2}\centering\footnotesize}

\title{HINT: Hierarchical Interaction Modeling \\for Autoregressive Multi-Human Motion \\Generation}

\author{Mengge Liu, Yan Di, Gu Wang, Yun Qu, Dekai Zhu, Yanyan Li, Xiangyang Ji \\
Tsinghua University\\
}

\iclrfinalcopy % Uncomment for camera-ready version, but NOT for submission.
\begin{document}

% \begin{iclrteaserfigure}
%   \includegraphics[width=\textwidth]{figs/MA.pdf}
%   \Description{Visualization of streaming multi-human generation.}
%   \label{fig:teaser_new}
% \end{iclrteaserfigure}

\maketitle

\thispagestyle{plain}
\pagestyle{plain}

\begin{figure}[H]
    \centering
    \includegraphics[width=0.9\linewidth]{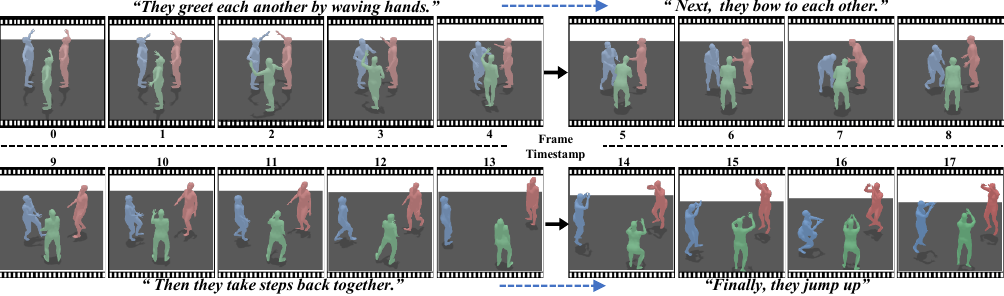}
    \caption{\textbf{Visualization of three-human motion generation results of HINT.} 
    By continuously updating the text guidance, HINT can autoregressively generate coherent, plausible human motions.}
    \label{fig:teaser_new}
\end{figure}

% \vspace{1em}

\input{sections/0_abstract}
\input{sections/1_intro}
\input{sections/2_related}
\input{sections/3_method}
\input{sections/4_exp}

\input{sections/9_conclusion}
% \input{sections/10_appendix}

\textbf{Ethics Statement.} 
This work includes a user study to evaluate the perceptual quality of generated motion sequences. 
All participants were adult volunteers who provided informed consent prior to participation. 
No personally identifiable or sensitive information was collected. 
The study was conducted in accordance with standard academic ethical practices, and participants were free to withdraw at any time without consequence.

\textbf{Reproducibility Statement.}
All experiments are conducted on publicly available datasets, and the implementation details, including model architectures, training procedures, and hyperparameters, are fully described in the main text and the appendix. 
Clear instructions for training and evaluation are provided in the Supplementary Material to ensure that the reported results can be reproduced under the same settings.
In addition, we will release the source code and pretrained models to facilitate verification and further research by the community.

\bibliography{ref}
\bibliographystyle{iclr2026_conference}

\input{sections/10_appendix}

\end{document}

%% file: math_commands.tex
\usepackage{amsmath,amsfonts,bm}

\def\eqref#1{equation~\ref{#1}}
\def\1{\bm{1}}

\DeclareMathAlphabet{\mathsfit}{\encodingdefault}{\sfdefault}{m}{sl}
\SetMathAlphabet{\mathsfit}{bold}{\encodingdefault}{\sfdefault}{bx}{n}

%% file: sections/0_abstract.tex
\begin{abstract}
Text-driven multi-human motion generation with complex interactions remains a challenging problem.
Despite progress in performance, existing offline methods that generate fixed-length motions with a fixed number of agents, are inherently limited in handling long or variable text, and varying agent counts.
These limitations naturally encourage autoregressive formulations, which predict future motions step by step conditioned on all past trajectories and current text guidance. 
%Such an online paradigm not only enables generating sequences of arbitrary length and arbitrary numbers of agents, but also allows for online editing of textual guidance to better model complex inter-person interactions.
In this work, we introduce \textbf{HINT}, the first autoregressive framework for multi-human motion generation with \textbf{H}ierarchical \textbf{INT}eraction modeling in diffusion. 
First, HINT leverages a disentangled motion representation within a canonicalized latent space, decoupling local motion semantics from inter-person interactions.
This design facilitates direct adaptation to varying numbers of human participants without requiring additional refinement. 
Second, HINT adopts a sliding-window strategy for efficient online generation, and aggregates local within-window and global cross-window conditions to capture past human history, inter-person dependencies, and align with text guidance.
This strategy not only enables fine-grained interaction modeling within each window but also preserves long-horizon coherence across all the long sequence. 
Extensive experiments on public benchmarks demonstrate that HINT matches the performance of strong offline models and surpasses autoregressive baselines. 
Notably, on InterHuman, HINT achieves an FID of 3.100, significantly improving over the previous state-of-the-art score of 5.154.

\end{abstract}

%% file: sections/1_intro.tex
\section{Introduction}

Human motion generation shows diverse applications spanning character animation~\citep{petrovich2022temos}, human-robot interaction~\citep{sahili2025text}, virtual reality~\citep{chen2024taming}, and content creation~\citep{tevethuman, guo2022generating}. 
Recently, text-driven approaches~\citep{javedintermask,liang2024intergen,zhaodartcontrol} have received growing attention, as they allow natural language to serve as a human-friendly control for generating semantically aligned human trajectories.
Beyond the single-human setting~\citep{tevethuman,zhang2024motiondiffuse,barquero2024seamless}, generating realistic, diverse, and controllable interactions for multiple humans remains highly challenging.

Existing approaches~\citep{javedintermask, liang2024intergen} are offline frameworks that generate motions of a fixed frame length and a fixed number of agents, as shown in Fig.~\ref{fig:teaser} (a). 
While effective for short sequences, these methods are inherently limited in handling variable-length natural language descriptions, dynamic interaction patterns, and varying agent counts. 
Moreover, they often fail to capture long-range dependencies across extended motion sequences, leading to incoherent or repetitive behaviors. These challenges naturally call for autoregressive formulations, as shown in Fig.~\ref{fig:teaser}~(b), where future motions are generated step by step conditioned on both past trajectories and textual instructions. 
Yet, autoregressive models in this domain remain underexplored, particularly for the multi-human setting with complex interactions.

In this work, we propose \textbf{HINT}, the first autoregressive diffusion-based framework for multi-human motion generation with hierarchical interaction modeling in diffusion. 
HINT presents two novel contributions. 
\textbf{First}, Canonicalized Latent Space, which encodes each human’s motion in its own local coordinate system, rather than encoding all agents in world coordinates~\citep{liang2024intergen}. 
Prior approaches~\citep{liang2024intergen, javedintermask, ruiz2024in2in} typically adopt such joint space, where motion dynamics are entangled with inter-agent positions, limiting scalability and requiring re-training when the number of agents changes. 
In contrast, HINT decouples individual motion representation from social interactions, while explicit relative transformations (rotations and translations) among agents are provided separately as conditions in diffusion. 
This separation enables the latent space to concentrate on motion semantics and ensures seamless adaptability to scenarios involving variable number of agents without finetuning or re-training.
\textbf{Second}, we propose a sliding-window strategy for efficient online generation.
Local to global temporal, spatial, and semantic cues are then aggregated to guide the diffusion process.
Local conditions are collected inside each window, \textit{i.e.}, target human's motion history, step index, partners' motion, and word-level text guidance, capturing fine-grained social and temporal dependencies within the window and preventing semantic drift.
Global conditions, including sequence index of the current window, total frame length, and compositional command text guidance, are used to locate the current window within the entire sequence and thereby enforce long-term consistency.
This hierarchical design enables natural, coherent, and semantically aligned multi-human motion generation, as shown in Fig.~\ref{fig:teaser_new}.

We conduct extensive experiments on the InterHuman~\citep{liang2024intergen} and InterX~\citep{xu2024inter} benchmarks. 
Results show that HINT not only matches the performance of strong offline methods but also outperforms existing autoregressive baselines by a large margin.

%% file: sections/2_related.tex
\begin{figure}[t]
\centering
    \centering
    \includegraphics[width=1.0\linewidth]{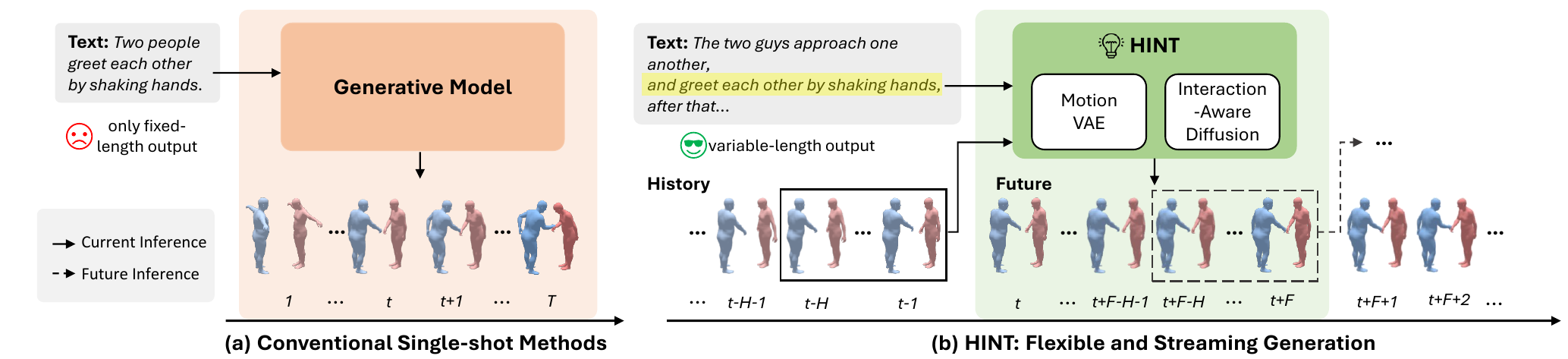}
    \caption{\textbf{Architecture Comparison.} (a) \textbf{Conventional Single-shot Methods}: Existing approaches (e.g., InterGen, in2IN, InterMask) generate motion sequences in a single shot with fixed length. (b)~\textbf{HINT:} Our framework integrates autoregressive and diffusion modeling to support streaming generation. Within a sliding window, the Interaction-Aware Diffusion leverages history and text to progressively synthesize future motions, thereby supporting open-ended, variable-length generation.}
    \label{fig:teaser}
\end{figure}
\section{Related Work}
\label{gen_inst}
\noindent\textbf{Single Human Motion Generation.} Recent work approaches single-person motion generation mainly with diffusion or autoregressive models.
Diffusion-based methods~\citep{tevethuman,chen2023executing,zhang2024motiondiffuse,barquero2024seamless} capture complex distributions and yield high-quality sequences across modalities such as text, audio, and scene context~\citep{tevethuman,xu2023interdiff,alexanderson2023listen}, but are typically limited to fixed-length clips.
Autoregressive models~\citep{jiang2023motiongpt,zhang2023generating} instead generate motions step by step, enabling variable-length synthesis and finer control, though they are prone to error accumulation.
DART~\citep{zhaodartcontrol} bridges these paradigms through a latent diffusion–autoregressive design that supports streaming, controllable motion generation.
We build on this paradigm to extend autoregressive diffusion to multi-human interaction, explicitly modeling semantic dependencies and coordination between participants.

\noindent\textbf{Human-Human Interaction Generation.} Recent years have witnessed increasing interest in human interaction motion generation~\citep{chopin2023interaction,xu2024regennet,liu2024physreaction,ghosh2024remos,tan2025think,liang2024intergen,shafir2024human,ruiz2024in2in,wang2024intercontrol,javedintermask,cai2024digital}, particularly in the areas of reaction and interaction generation.
Reaction generation aims to synthesize plausible responses conditioned on a partner’s motion, with approaches ranging from Transformer-based coordination~\citep{chopin2023interaction} and diffusion with distance constraints~\citep{xu2024regennet} to physics-driven modeling~\citep{liu2024physreaction}, spatio-temporal cross-attention~\citep{ghosh2024remos}, and reasoning with LLMs~\citep{tan2025think}.
Interaction generation instead models both humans jointly, using dual-branch diffusion~\citep{liang2024intergen}, lightweight communication across pretrained models~\citep{shafir2024human}, dual-level textual prompts~\citep{ruiz2024in2in}, LLM-based planning~\citep{wang2024intercontrol}, or masked spatio-temporal token prediction~\citep{javedintermask}.
Beyond pairwise interactions, SocialGen~\citep{yu2025socialgen} leverages language models for group social behaviors, Multi-Person Interaction Generation~\citep{xu2025multi} scales two-person priors to larger groups, and PINO~\citep{ota2025pino} enables long-duration and customizable generation for arbitrary group sizes.
Despite these advances, most methods still generate fixed-length sequences, limiting their applicability in real-time and streaming scenarios.

%% file: sections/3_method.tex
\section{Method}
\label{sec:method}

\begin{figure}[t]
\centering
    \centering
    \includegraphics[width=1.0\linewidth]{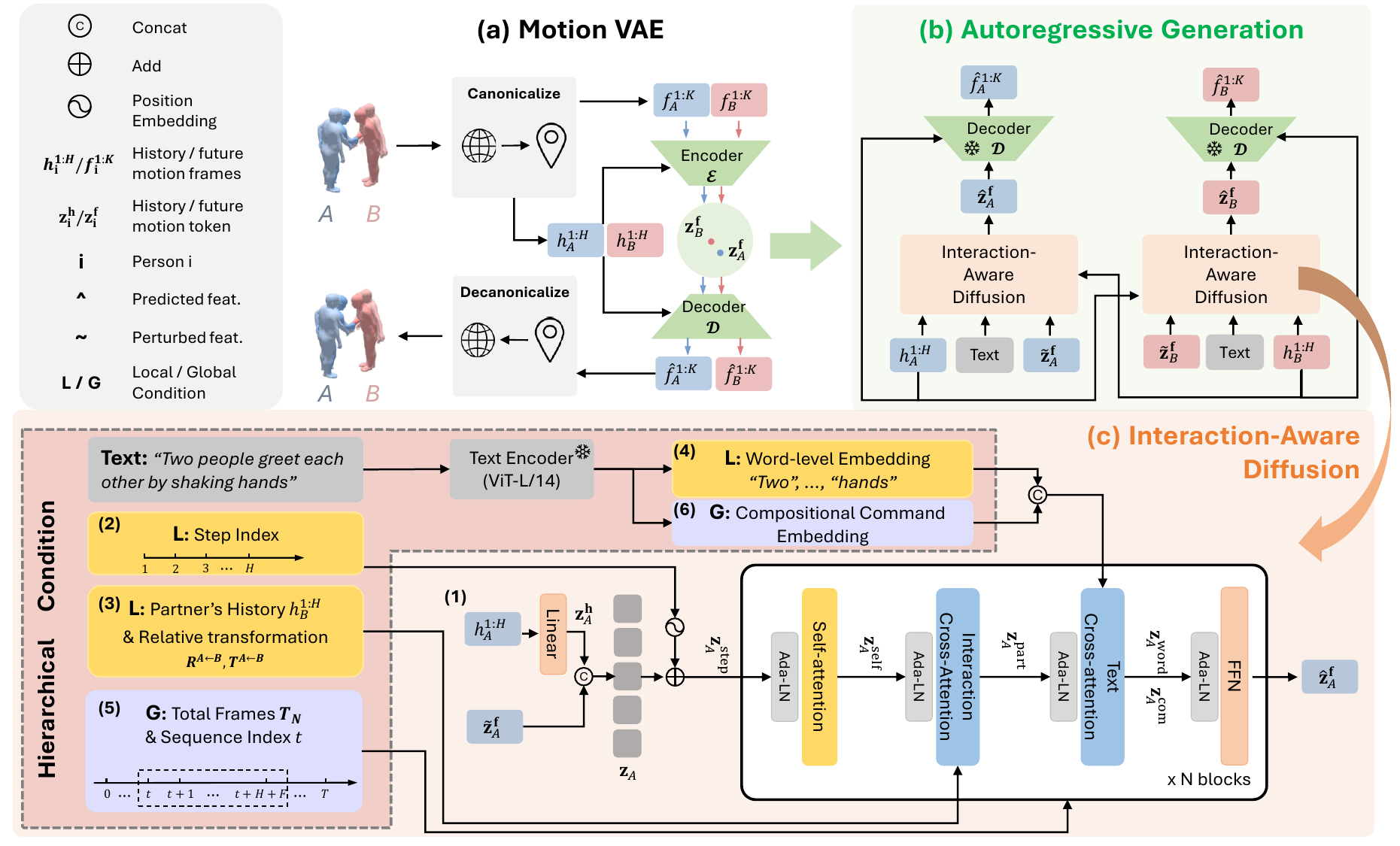}
    \caption{\textbf{Overview of HINT in two-human interaction generation.} (a) Canonicalized latent space. (b) Within this latent space, motion is generated in a sliding-window autoregressive manner, where the Interaction-Aware Diffusion predicts the next $K$ frames. (c) The detailed architecture of the Interaction-Aware Diffusion, in which hierarchical conditions guide the generation process. }
    \label{fig:method}
\vspace{-0.2cm}
\end{figure}

% \textbf{Objective.} 
We address text-driven online multi-human motion generation, which sequentially predicts 
% a set of human motion trajectories, where the 
future poses of $N$ agents 
% are generated 
conditioned on their past motions and a textual description $\mathcal{T}$. 
Formally, let  
\begin{equation}
\label{eq:motion_seq}
\mathbf{M}^{1:T} = \left\{ \mathbf{m}_{(i)}^{t} \in \mathbb{R}^d \;\middle|\; i = 1,\dots,N;\; t=1,\dots,T \right\}
\end{equation}
denote the motion sequence of $N$ humans over $T$ timesteps, where $\mathbf{m}_{(i)}^{t}$ is the motion representation of agent $i$ at time $t$,
$d$ is the dimension of the representation.
Autoregressive multi-human motion generation recursively predicts  
\begin{equation}
\hat{\mathbf{M}}^{t:t+K} \sim p_\theta\!\left(\mathbf{M}^{t:t+K} \;\middle|\; \hat{\mathbf{M}}^{1:t-1}, \mathcal{T}^{1:t+K}\right),
\end{equation}
with trained parameters $\theta$, thereby capturing both temporal dependencies across timesteps and social dependencies across humans.
We jointly predict $K$ future timesteps for efficiency.
% \iffalse
% Given an interaction text $\mathcal{T}$ and the initial motions of $N$ individuals $H=[H^1,H^2,\ldots,H^N]$, the goal is to autoregressively generate their future motions $F=[F^1,F^2,\ldots,F^N]$. 
% \fi
% \iffalse
% The generation process is divided into $T$ rollouts, each of length $K$:
% $$
% F^i = [R_1^i, R_2^i, \ldots, R_T^i],  i=1,\ldots,N\quad
% R_t = \{R_t^1, R_t^2, \ldots, R_t^N\}, \quad R_t^i \in \mathbb{R}^{K \times D},
% $$
% where $R_t^i$ denotes the $K$ frames generated for the $i$-th person in the $t$-th window, and $D$ is the motion dimension.
% \fi
For clarity, we use $h_{A}^{1:H}$ to represent the $H$-timestep history motion of agent $A$ and $f_{A}^{1:K}$ to represent the $K$-timestep future motion within a sliding window, as shown in Fig.~\ref{fig:method}. 
We empirically set $H=4, K=16$. 
%We primarily focus on two-human interaction generation as the core scenario, while demonstrating that our framework naturally generalizes to multi-human settings.

\subsection{Overview of HINT}
Fig.~\ref{fig:method} demonstrates the \textbf{two-human} motion generation pipeline of \textbf{HINT}, 
which employs a sliding-window strategy to autoregressively extend future segments with a diffusion model (see Fig.~\ref{fig:teaser}).
We show that HINT naturally generalizes to \textbf{multi-human} settings in Sec.~\ref{multi_human}
%\textcolor{red}{This content is incomplete, only introducing the encoder. What about the decoder? What does the decoder output? What is the shape of the latents encoded by the encoder? Also, you should mention that the encoder and decoder are frozen and do not require retraining in subsequent stages.}

\textbf{Motion VAE.} 
In Fig.~\ref{fig:method} (a), we first construct a canonicalized shared latent space to map raw motion sequences into latent representations. 
Concretely, the motion of each individual, $m_A, m_B \in \mathbb{R}^{T \times d}$, with $T$ timesteps and $d$ dimensions, is divided into overlapping windows and canonicalized in its local coordinate to remove absolute position. 
A transformer-based Motion VAE, following DART~\citep{zhaodartcontrol}, is then employed for sliding-window modeling. 
The VAE consists of an encoder $\mathcal{E}$ and a decoder $\mathcal{D}$. 
Given a window with $H$ history $h_{(i)}^{1:H}$ and $K$ future frames $f_{(i)}^{1:K}$ for human $i\in\{A,B\}$, $\mathcal{E}$ conditions on the history and encodes the future into a latent vector $\mathbf{z}_{(i)}^{\mathbf{f}} \in \mathbb{R}^{l}$ in the shared latent space, where $\mathbf{z}_{(i)}^{\mathbf{f}}$ denotes the future motion representation of agent $i$, and $l$ is the latent dimensionality. 
 $\mathcal{D}$ reconstructs the $K$ future frames from $\mathbf{z}_{(i)}^{\mathbf{f}}$, conditioned on the corresponding history.
%The VAE is first trained independently in the first stage. 
Once trained, both encoder $\mathcal{E}$ and decoder $\mathcal{D}$ are frozen in subsequent modules, ensuring that the latent space remains stable and consistent.  
%This shared latent space enables natural scalability to multi-person scenarios and provides stable and compact representations for downstream generation.

\textbf{Sliding-Window Strategy for Autoregressive Generation.} 
As illustrated in Fig.~\ref{fig:method}~(b), HINT employs a sliding-window autoregressive process for both training and generation. 
During training, the ground-truth motion sequence is divided into overlapping windows. 
Future frames in each window are encoded into a latent $\mathbf{z}_{(i)}^{{\mathbf{f}}}, i \in \{A,B\}$ by the motion encoder $\mathcal{E}$, where noise is added to obtain $\tilde{\mathbf{z}}_{(i)}^{{\mathbf{f}}}$. 
Conditioned on $\tilde{\mathbf{z}}_{(i)}^{{\mathbf{f}}}$, historical motion $h_{(i)}^{1:H}$, and text $\mathcal{T}$, the Interaction-Aware Diffusion model learns to predict the denoised latent $\mathbf{\hat{z}}_{(i)}^{\mathbf{f}}$. 
%The frozen decoder then reconstructs the future $K$ frames $\hat{f}_{(i)}^{1:K}$. 
%The predicted frames are appended to the history and used as conditions for subsequent windows. 
During inference, we sample a future latent $\mathbf{\hat{z}}_{(i)}^{\mathbf{f}}$ conditioned on the history and text.
Then the frozen decoder $\mathcal{D}$ reconstructs the future $K$ frames $\hat{f}_{(i)}^{1:K}$ from $\mathbf{\hat{z}}_{(i)}^{\mathbf{f}}$.
$\hat{f}_{(i)}^{1:K}$ is appended to the history to condition the next window, proceeding autoregressively.

%\textcolor{red}{This section is incomplete. According to Section 3.3 below, you need to specify the names of various conditioning inputs and describe how they are fed into the network. While you don't need to provide extremely detailed architecture, you should outline the general structure of the network, including its inputs and outputs. All the symbols shown in your figure should be mentioned and explained as comprehensively as possible. Additionally, modules such as the transfer module must be described in terms of their functionality. Furthermore, the method for encoding text should be explained in detail.}

\textbf{Interaction-Aware Diffusion.} 
Fig.~\ref{fig:method} (c) illustrates the Interaction-Aware Diffusion module, exemplified with two-human interaction generation.
The model employs shared weights and shared conditioning signals for both agents $A$ and $B$, ensuring that the same model parameters are applied symmetrically across the two. 
% For clarity, we describe the pipeline from the perspective of
Taking the motion generation of human $A$ as an example, the model is first conditioned on the historical motion sequences of both human $A$ and human $B$, denoted as $h_A^{1:H}$ and $h_B^{1:H}$, respectively. 
Then the relative rotation $\mathbf{R}^{A \leftarrow B}$ and translation $\mathbf{T}^{A \leftarrow B}$ are encoded and integrated.
Additionally, the module integrates a set of hierarchical conditions: 1) step index within the current temporal window; 2) word-level text embedding; 3) sequence index of the current window in the overall sequence, total frame length $T_N$ to be generated according to the textual description; and 4) compositional command embedding.
Together, these conditions encode both textual semantics and multi-scale positional information, enabling the model to capture fine-grained temporal dependencies while maintaining global consistency across the generated motion sequence.

%In Fig.~\ref{fig:method} (c), we use a latent diffusion model to predict future motion sequences. The module conditions on history motion $H_A, H_B$, interaction text $\mathcal{T}$, and relative positions between agents $R^{A \rightarrow B}, T^{A \rightarrow B}$, thereby explicitly modeling individual dynamics, cross-person dependencies, and semantic alignment. To align long text and motion sequences, we introduce a hierarchical motion condition mechanism that aligns semantic units with motion segments. %Consequently, HINT ensures long-sequence coherence and scalability through autoregressive sliding windows, while achieving high-fidelity, temporally smooth motion generation via diffusion refinement.

% \textcolor{red}{Introduce the overall inference procedures here. How to use the sliding window strategy to generate future motions.}

\subsection{Canonicalized Latent Space}
\label{cl_space}
We propose a \textit{Canonicalized Latent Space}, as illustrated in Fig.~\ref{fig:method} (a), which encodes explicit motion sequences into compact latents.
Existing methods~\citep{liang2024intergen, javedintermask} on human-human interaction generation often adopt a Joint Multi-Human Latent Space by applying the same coordinate transformation to both agents' motions, which preserves relative position
% and encoding their motions respectively. 
but entangles motion semantics with inter-person geometry, hindering generalization to multi-human scenarios.
Instead, we perform independent canonicalization and normalization for each agent $A, B$, transforming motions to their respective local coordinate, while explicitly encoding relative transformations (rotation $\mathbf{R}^{A \leftarrow B}$ and translation $\mathbf{T}^{A \leftarrow B}$) as conditions for the generative model. 
Specifically, after canonicalization, each individual is oriented toward the positive $z$-axis, with the root joint positioned at the origin. 
%This decoupling fundamentally reduces semantic–geometric entanglement and improves scalability.

Formally, given a motion sequence $\mathbf{M}_{(i)}$ for human $i \in \{A,B\}$, the canonicalized motion is defined as $\mathbf{M}_{(i)}^{\mathbf{c}} = \mathbf{R}_{(i)} \mathbf{M}_{(i)} + \mathbf{T}_{(i)}$, where $\mathbf{R}_{(i)}$ and $\mathbf{T}_{(i)}$ denote the rotation and the translation, respectively.
Following the reparameterization strategy~\citep{kingma2013auto}, the latent representation $\mathbf{z}_{(i)}^\mathbf{f}$ is obtained via the encoder $\mathcal{E}$ as
\begin{equation}
\mathbf{z}_{(i)}^\mathbf{f} \sim q_\phi(\mathbf{z}_{(i)}^\mathbf{f} \mid \mathbf{M}_{(i)}^{\mathbf{c}}),
\end{equation}
where $q_\phi$ is a Gaussian inference network. Training objectives are described in Sec.~\ref{sec:Training Strategy}.
To inject relative positional information, we compute the relative rigid transformation as follows,
\begin{equation}
\mathbf{R}^{i \leftarrow j} = \mathbf{R}_{(i)}\mathbf{R}_{(j)}^{\top}, \quad
\mathbf{T}^{i \leftarrow j} = \mathbf{T}_{(i)} - \mathbf{R}^{i \leftarrow j}\mathbf{T}_{(j)}.
\end{equation}
% for human $i\in {A,B}$, the latent code is obtained as:
% $$
% \mathbf{z}_{(i)} \sim q_\phi(\mathbf{z}_{(i)} \mid \mathbf{M}_{(i)}^c), \quad 
% \hat{\mathbf{M}}_{(i)} = \text{Transf}(D_\theta(\mathbf{z}^{(i)}, \mathbf{H}^{(i)}),R_{(i)},T_{(i)}),
% $$
% where $q_\phi$ is a Gaussian inference network, $\mathbf{H}^{(i)}$ denotes history frames, and $D_\theta$ is the decoder, Transf is the Coordinate Transformation module. Training objectives are described in Sec.~3.4. 

% \textcolor{red}{Canonicalized Latent Space vs. Joint multi-human latent space. Here you can introduce the advantage of this canonicalized embedding over traditional encoding} 
This transformation $[\mathbf{R}^{i \leftarrow j}, \mathbf{T}^{i \leftarrow j}]$ is encoded into the diffusion network as a condition term.

\textbf{Canonicalized Latent Space \textit{vs.} Joint Multi-Human Latent Space}.
Our canonicalized latent space has two advantages over previous joint multi-human latent space~\citep{liang2024intergen, javedintermask,ruiz2024in2in}.
First, it effectively disentangles absolute position information from motion dynamics, forcing the latent to focus on the movement patterns themselves without being biased by spatial location.
Second, such design enforces cross-human consistency in the latent space, thereby facilitating robust generalization to interactions involving three or more humans.

\subsection{Hierarchical Motion Condition}
\label{hmc}
To enable effective autoregressive motion generation, we incorporate local-to-global guidance into the diffusion process. 
Built upon a latent diffusion backbone, our \textit{Hierarchical Motion Condition} (HMC) strategy organizes temporal, spatial, and semantic cues into multi-level conditions.
We provide local conditions, which capture short-term dependencies and fine-grained semantic alignment within the current window, and global conditions, which enforce long-term consistency across the sequence described by the text guidance.
We demonstrate HMC on human-human interaction generation (Fig.~\ref{fig:method} (c)), illustrating the procedure from human $A$'s perspective, as weights and conditions are shared across both humans. 
% As the weights and conditions are shared for both humans, we illustrate the process using human $A$ for clarity.
%\textcolor{red}{Should refer to the figure}

\textbf{Local Conditions.}
Within each window, we employ four types of local conditions as follows.
%the model is conditioned on four types of local information as follows.

\textit{1) Target Human History Embedding.} 
As shown in Fig.~\ref{fig:method} (c-1), the history motion of human $A$, $h_A^{1:H}$, is first mapped into a feature representation $\mathbf{z}_A^\mathbf{h}$ via a linear projection. $\mathbf{z}_A^\mathbf{h}$ is then concatenated with the future motion token $\mathbf{z}_A^\mathbf{f}$ to form the motion feature $\mathbf{z}_A$.
% Human A's motion feature merged with step index $\mathbf{z}_A^{\text{step}}$, is further processed via self-attention to capture temporal dependencies on history motion:
% $$
% \mathbf{z}_{A}^{\text{self}} = \mathrm{Attn}(\mathbf{z}_A^{\text{step}}, \mathbf{z}_A^{\text{step}}, \mathbf{z}_A^{\text{step}}).
% $$

\textit{2) Step Index.} 
Both human $A$ and $B$ provide $H$ history frames. Each history frame is indexed by its timestep from 
% $0$ to $H-1$, 
{$1$ to $H$},
and the index is encoded into an embedding $\mathbf{e}_s$ (Fig.~\ref{fig:method} (c-2)). Adding this to the motion feature $\mathbf{z}_A$ yields
$
\mathbf{z}_A^{\text{step}} = \mathbf{z}_A + \mathbf{e}_s,
$
 which is then processed via self-attention to capture temporal dependencies:
\begin{equation}
\mathbf{z}_{A}^{\text{self}} = \mathrm{SelfAttn}(\mathbf{z}_A^{\text{step}}, \mathbf{z}_A^{\text{step}}, \mathbf{z}_A^{\text{step}}).
\end{equation}
This enables the model to reason about temporal ordering within the prediction window. 

\textit{3) Partner History Embedding.} 
To model interactions, human $B$'s history $h_B^{1:H}$ is transformed into human $A$'s local coordinate via relative rotation $\mathbf{R}^{A\leftarrow B}$ and translation $\mathbf{T}^{A\leftarrow B}$ (Fig.~\ref{fig:method} (c-3)):
\begin{equation}
h_{B \rightarrow A}^{1:H} = \mathbf{R}^{A \leftarrow B} h_B^{1:H} + \mathbf{T}^{A \leftarrow B}.
\end{equation}
This is integrated into the diffusion model through Interaction Cross-Attention:
\begin{equation}
\mathbf{z}_{A}^{\text{part}} = \mathrm{CrossAttn}(\mathbf{z}_{A}^{\text{self}}, h_{B \rightarrow A}^{1:H}, h_{B \rightarrow A}^{1:H}).
\end{equation}
\textit{4) Word-Level Text Embedding.} 
Finally, we introduce word-level text embedding to impose fine-grained semantic constraints within each window, as depicted in Fig.~\ref{fig:method} (c-4). 
Since a single sentence may be very long or contain complex commands, we split it into words, each serving as a token, $\mathbf{E}_{\text{word}} = [\mathbf{e}_1,\dots,\mathbf{e}_L]$, where $\mathbf{e}_l$ denotes the embedding of the $l$-th token, and integrated into latent features via Text Cross-attention:
$$
\mathbf{z}_A^{\text{word}} = \mathrm{CrossAttn}(\mathbf{z}_{A}^{\text{part}}, \mathbf{E}_{\text{word}}, \mathbf{E}_{\text{word}}).
$$
By jointly leveraging individual history, step index, partner history, and token-level text embedding, the model captures fine-grained interaction patterns and achieves precise text–motion alignment within each rollout, thereby alleviating semantic drift in long-sequence generation.

\textbf{Global Conditions.}
Across all windows, we collect the following two types of global information.

\textit{1) Sequence Index and Total Frame Number.} 
In Fig.~\ref{fig:method} (c-5), we first incorporate both the global sequence index $t$ and the total number of frames $T_N$ of the corresponding text segment, which indicate the position of the current window within the entire motion sequence and the overall sequence length, respectively. 
This information is then injected into the diffusion network through Adaptive Layer Normalization (AdaLN) \citep{peebles2023scalable}, ensuring that the generation process is aware of both the frame-level position and the global temporal context.

\textit{2) Compositional Command Embedding.} 
If the user provides a textual description for the entire sequence, where the description consists of multiple interconnected commands that drive the human body to achieve one or more specific goals, we encode the whole text $\mathcal{T}$ into a single global token $\mathbf{e}$ to serve as guidance for the sequence generation, as depicted in Fig.~\ref{fig:method} (c-6). 
Then $\mathbf{e}$ is injected into the model through Text Cross-Attention to provide global semantic guidance across windows:
\begin{equation}
\mathbf{z}_{A}^{\text{com}} = \mathrm{CrossAttn}(\mathbf{z}_{A}^{\text{part}}, \mathbf{e}, \mathbf{e}).
\end{equation}
Conceptually, word-level embeddings $\mathbf{E}_{\text{word}} = [\mathbf{e}_1,\dots,\mathbf{e}_L]$ serve as local conditions while compositional command embedding $\mathbf{e}$ functions as the global condition. 
In practice, however, both are concatenated and jointly fed into the same Text Cross-Attention block, allowing simultaneous modeling of fine-grained semantics and holistic context within a unified interaction.

% This hierarchical conditioning 
%HMC provides latent diffusion with more expressive local constraints and stronger global consistency, ensuring natural, coordinated, and semantically aligned human-human interaction generation.

\subsection{Training Strategy}
\label{sec:Training Strategy}

We adopt a two-stage training strategy that decouples motion encoding and generation. 
% We first pretrain the MVAE to learn stable latent representations, and then train the interaction-aware diffusion while keeping the MVAE frozen.

\textbf{Stage I: Motion VAE Pretraining.}
The Motion VAE is pretrained to obtain stable latent representations by 
% In the first stage, we optimize 
optimizing the standard VAE objective~\citep{kingma2013auto}:
% , including a reconstruction loss and a KL divergence term:
\begin{equation}
\mathcal{L}_{\text{VAE}} 
= \sum_i \mathcal{L}_{\mathrm{rec}}(\hat{\mathbf{M}}_{(i)}, \mathbf{M}_{(i)}) 
+ \beta \, \mathcal{L}_{\mathrm{KL}}\big(q_\phi(\mathbf{z}_{(i)}^\mathbf{f} \mid \mathbf{M}_{(i)}^{\mathbf{c}})\,\|\,p(\mathbf{z}_{(i)}^{\mathbf{f}})\big),
\end{equation}
where $\mathcal{L}_{\mathrm{rec}}$ reconstructs future frames, $\mathcal{L}_{\mathrm{KL}}$ regularizes the latent distribution with KL divergence, $\beta$ is a balancing factor, and $p(\mathbf{z}_{(i)}^{\mathbf{f}})$ denotes the standard Gaussian prior. 
After pretraining, the encoder and decoder are frozen in generation.

% \textcolor{red}{This equation is wrong. Use the symbols above, there is no X in your pipeline. Where is q?}

\textbf{Stage II: Diffusion with Autoregressive Sliding Window.}
We train the Interaction-Aware Diffusion model using an autoregressive sliding-window strategy. 
At each window, the diffusion model is optimized with the standard denoising loss $\mathcal{L}_{\text{diff}}$, augmented by  interaction-specific regularizers inspired by InterGen~\citep{liang2024intergen}, including joint affinity $\mathcal{L}_{\text{aff}}$, cross-person distance constraint $\mathcal{L}_{\text{dist}}$, and relative orientation constraint $\mathcal{L}_{\text{ori}}$:
\begin{equation}
\mathcal{L}
= \mathcal{L}_{\text{diff}}
+ \lambda_{\text{aff}}\mathcal{L}_{\text{aff}}
+ \lambda_{\text{dist}}\mathcal{L}_{\text{dist}}
+ \lambda_{\text{ori}}\mathcal{L}_{\text{ori}},
\end{equation}
where $\lambda_\ast$ indicates balancing weights. 
Please refer to Appendix \ref{App:training details} for details of each loss term.
%This two-stage strategy 
% effectively separates representation learning from generative modeling, 
%ensures stable latent representations while improving the fidelity, coherence, and naturalness of generated human-human interactions.

\subsection{From Two-Human to Multi-Human Motion Generation}
\label{multi_human}
Built upon the Canonicalized Latent Space (Sec.~\ref{cl_space}) and the shared-weight Interaction-Aware Diffusion model (Fig.~\ref{fig:method} (c)), HINT naturally generalizes to multi-human interaction scenarios. 
The proposed space can be directly applied to multi-human motion generation without additional training, since it decouples individual motion with social interactions.
For the diffusion model, we only update one condition term: partner history embedding (Sec.~\ref{hmc}, Local Conditions) by directly concatenating all partners motion history and feeding them into the diffusion.
We do not perform any fine-tuning when scaling to more humans.
Here, we only provide the most straightforward extension from two-person to multi-person motion generation.
If additional multi-person motion datasets are employed, fine-tuning the cross-attention module is expected to yield further performance improvements.
Please refer to Supplementary Materials for video results.

%% file: sections/4_exp.tex
\input{tables/results}
\section{Experiments}
\label{sec:exp}

% \subsection{Experiment Setup}
\textbf{Datasets.} 
We evaluate HINT on InterHuman~\citep{liang2024intergen} and InterX~\citep{xu2024inter}. 
\textit{InterHuman} comprises 7,779 motion sequences paired with 23,337 unique textual annotations containing 5,656 distinct words.
It is built upon the SMPL-H body model, and we adopt a motion representation similar to InterGen~\citep{liang2024intergen}, where each frame is expressed as $x^i = [\mathbf{j}_l^p, \mathbf{j}_l^v, \mathbf{j}^r, \mathbf{c}^f]$. Here, $\mathbf{j}_l^p \in \mathbb{R}^{3N_j}$ and $\mathbf{j}_l^v \in \mathbb{R}^{3N_j}$ represent the joint positions and velocities in the normalized local frame, $\mathbf{j}^r \in \mathbb{R}^{6(N_j-1)}$ denotes the 6D rotation \citep{rot6d_zhou2019} of each joint in the root frame, $\mathbf{c}^f \in \mathbb{R}^4$ is a binary foot-ground contact feature, and $N_j$ denotes the number of joints, set to $22$ for InterHuman. 
\textit{InterX} is based on the SMPL-X body model and contains 13,888 motion sequences with 34,164 fine-grained textual descriptions. 
Each motion frame is represented as $x^i = [\mathbf{j}^r, \mathbf{r}_l^p, \mathbf{r}_l^v]$, where $\mathbf{j}^r \in \mathbb{R}^{6N_j}$ is the 6D rotation in the normalized local frame, and $\mathbf{r}_l^p \in \mathbb{R}^3$, $\mathbf{r}_l^v \in \mathbb{R}^3$ denote the root joint’s position and velocity in the local frame, respectively. 
For InterX, $N_j$ is set to $55$.

\textbf{Baselines.} 
Offline baselines: T2M \citep{guo2022generating}, MDM \citep{tevethuman}, ComMDM \citep{shafir2024human}, InterGen \citep{liang2024intergen}, MoMat-MoGen \citep{cai2024digital}, in2IN \citep{ruiz2024in2in}, and InterMask \citep{javedintermask} on InterHuman, and with T2M, MDM, ComMDM, InterGen, and InterMask on InterX.  
In addition, we introduce two extended baselines: InterMask*, which denotes our online adaptation of InterMask, and DART$^{\dagger}$, which denotes our extension of DART \citep{zhaodartcontrol} from single-human to two-human scenarios(see Appendix~\ref{appendix:baseline_details} for details).
% Detailed descriptions and implementation settings are in Appendix~\ref{appendix:baseline_details}.

\textbf{Evaluation Metrics.} 
%Following InterGen \citep{liang2024intergen}, we employ several metrics to evaluate the generated motions. 
%\textit{R-Precision} (reported as R@Top3; see Appendix~\ref{addition results} for Top1/Top2) and \textit{Multimodal Distance} (MM Dist) assess text-motion consistency by measuring, respectively, the ranking of Euclidean distances between motion and text embeddings, and the average Euclidean distance between each generated motion and its corresponding text. 
\textit{R-Precision} (reported as \textit{R@Top3}; see Appendix~\ref{addition results} for \textit{R@Top1/Top2}) and \textit{Multimodal Distance} (\textit{MM Dist}) are used to evaluate text-motion consistency. 
Specifically, \textit{R-Precision} measures the rank of the Euclidean distance between motion and text embeddings, while \textit{MM Dist} computes the average Euclidean distance between each generated motion and its corresponding text.
\textit{Frechet Inception Distance} (\textit{FID}) evaluates the similarity in the feature space between generated and ground-truth motions, reflecting motion realism. 
\textit{Diversity} (\textit{Div}) measures motion variety via average pairwise feature distances among generated motions. 
All methods are evaluated 20 times with different random seeds, and we report the mean results with the $95\%$ confidence interval.

\textbf{Inference Speed.} HINT takes about 1.1s to generate 16 future frames from a single window on a single NVIDIA GeForce 3090 GPU, while DART{$^\dagger$} takes about 0.3s and InterMask* takes about 1.1s under the same conditions.

\subsection{Comparison with Baselines}

\begin{wrapfigure}[10]{R}{0.45\textwidth}
    \centering
    \vspace{-1.0cm}
    \includegraphics[width=\linewidth]{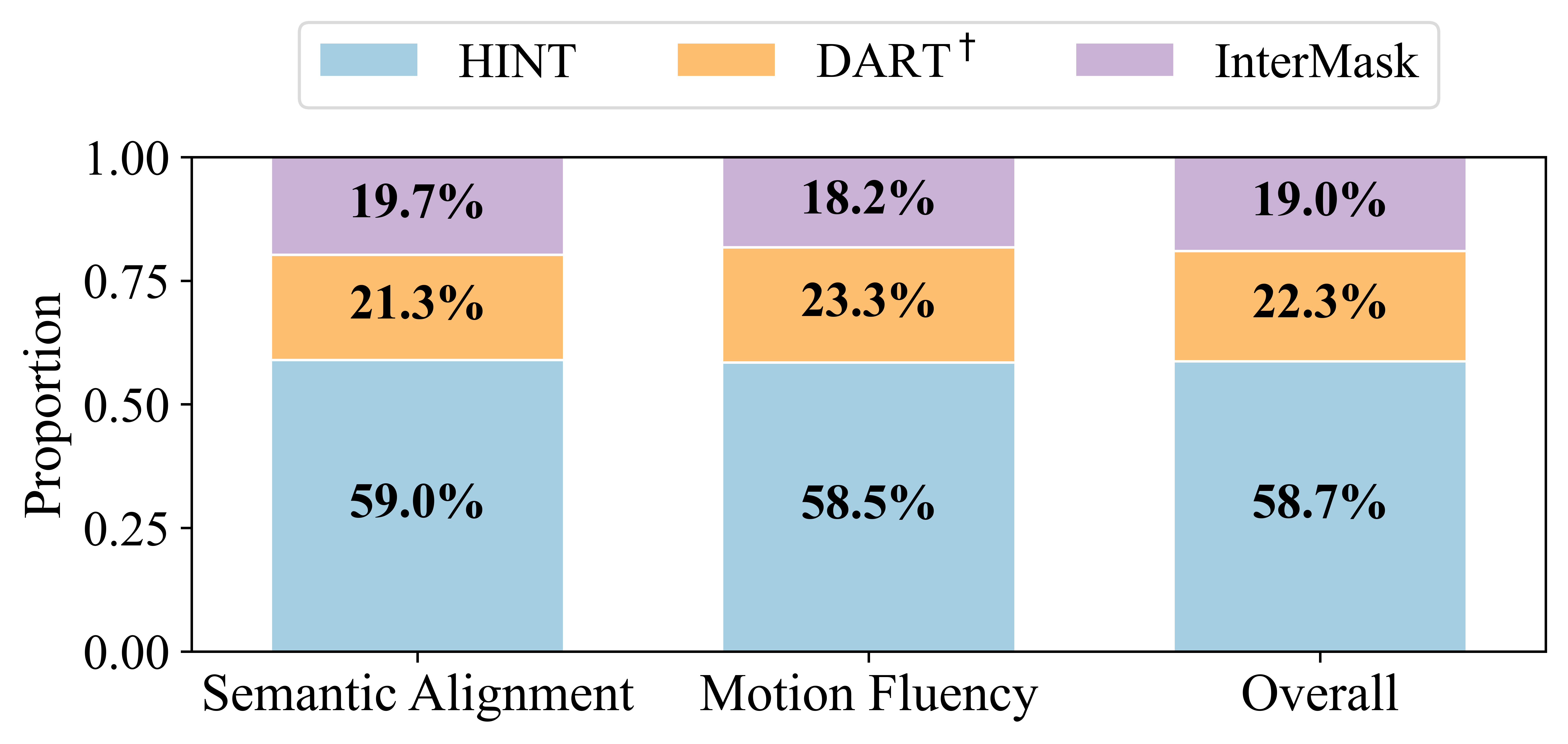}
    % \vspace{0.5cm}
    \caption{ User study between HINT, DART$^{\dagger}$ and InterMask.}
    \label{fig:user_study}
\end{wrapfigure}

Tab.~\ref{tab:quantitative_interhuman_interx_combined} presents the evaluation results.
Among all compared methods, HINT achieves state-of-the-art FID scores of 3.100 on InterHuman and 0.278 on InterX, improving 2.054 and 0.121 over the second-best method, InterMask. 
This significant gain highlights the superior realism and naturalness of the motions generated by HINT, which can be primarily attributed to HINT’s hierarchical interaction modeling strategy.
It explicitly and comprehensively conditions on past motion histories and relative position relations between humans. 
As a result, within each sliding window, HINT is able to effectively capture and construct rich inter-human interactions.
For other metrics, HINT is consistently superior to online competitors InterMask* and DART$^{\dagger}$, while slightly inferior to the offline method InterMask. 
For instance, on InterHuman, compared to InterMask, HINT shows a small decrease of 0.011 in \textit{R@Top3} and 0.006 in \textit{MM Dist}.
As an autoregressive method, HINT does not perform global optimization, which inevitably leads to insufficient alignment with the global text command.
Overall, these experiment results validate HINT's effectiveness.

% \begin{wrapfigure}[10]{R}{0.45\textwidth}
%     \centering
%     \vspace{-0.7cm}
%     \includegraphics[width=\linewidth]{figs/user_study.png}
%     %\vspace{-0.7cm}
%     \caption{ User study between HINT, DART$^{\dagger}$ and InterMask.}
%     \label{fig:user_study}
% \end{wrapfigure}

\begin{figure}[t]
\centering
    \centering
    \includegraphics[width=0.98\linewidth]{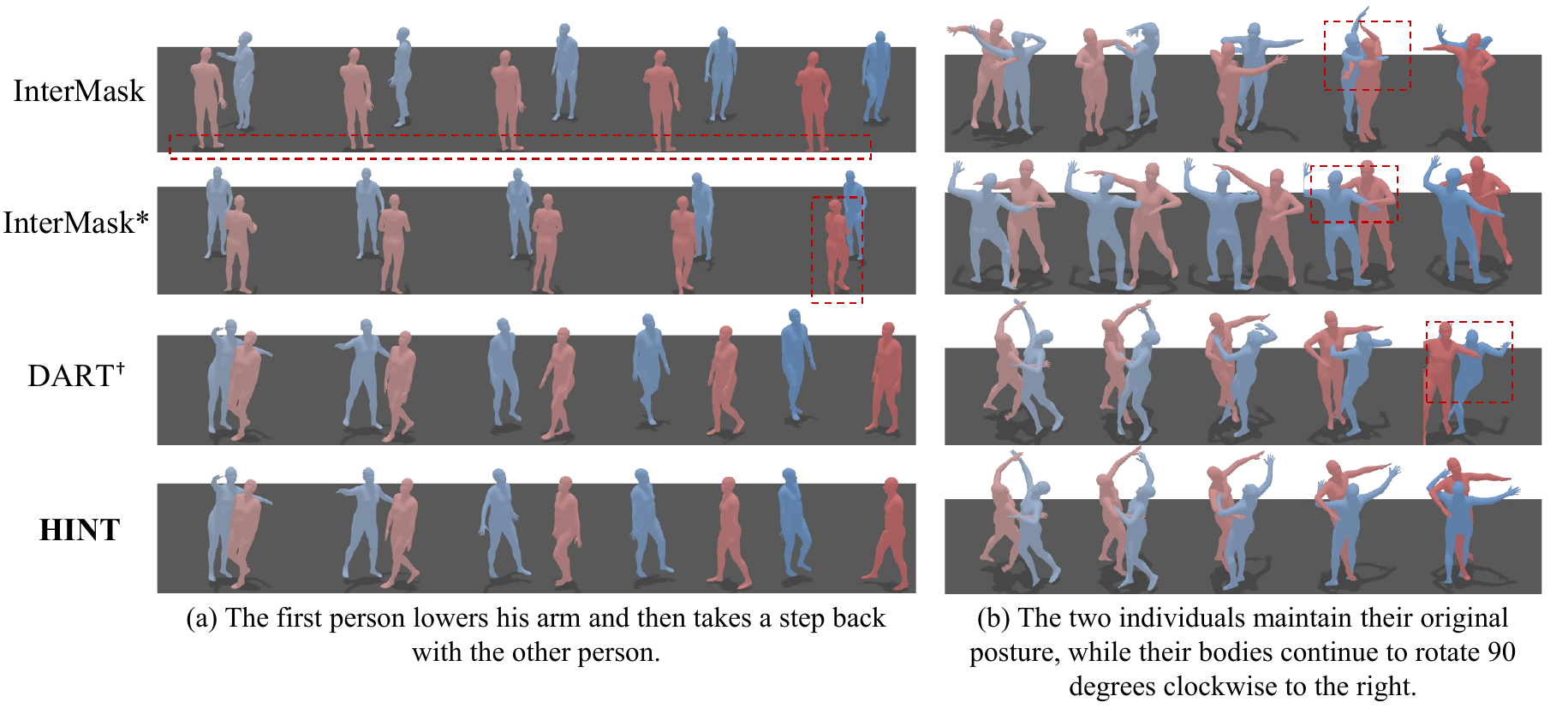}
    \caption{Visual comparisons of InterMask, InterMask*, DART$^{\dagger}$ and HINT on InterHuman. 
    HINT performs better in regions with complex interactions.}
    \label{fig:exp_fig1}
    \vspace{-0.5cm}
\end{figure}

Fig.~\ref{fig:exp_fig1} shows qualitative comparisons of InterMask, InterMask*, DART$^{\dagger}$, and HINT trained on InterHuman with the same text descriptions. 
In (a), InterMask fails to generate the backward motion of the left person, InterMask* produces less natural movements, while both DART$^{\dagger}$ and HINT align well with the text. 
In (b), InterMask and InterMask* fail to generate motions consistent with the semantics, DART$^{\dagger}$ does not explicitly model interactions and thus shows weaker interaction quality, whereas HINT achieves superior semantic alignment, interaction effectiveness, and motion fluency. 
More visual results are provided in Appendix \ref{addition results}.
Videos are provided in Supplementary Materials.

\noindent\textbf{User Study.}
To further evaluate the subjective quality of the generated results, we conducted a user study.
Fifteen participants are invited to compare HINT against InterMask and DART$^{\dagger}$ in terms of semantic alignment and motion fluency.
Results are shown in Fig. \ref{fig:user_study}.
HINT received over $50\%$ of the votes across all metrics.

\subsection{Ablation Studies}

\input{tables/abla_ih_key_components}

%Tabs.~\ref{tab:ablation_studies_MVAE} and \ref{tab:ablation_studies} ablate HINT's key components on InterHuman.
% dataset to evaluate the contribution of the key components in HINT, as shown in .

% This forces the model to encode pose details and spatial relations simultaneously, weakening its ability to capture motion dynamics and degrading reconstruction quality. 
% In contrast, our Canonicalized Latent Space disentangles positional information from motion details via independent normalization, allowing the model to focus on motion itself and achieve superior reconstruction.

\input{tables/abla_ih_cls} 

Tab.~\ref{tab:ablation_studies_MVAE} further compares our Canonicalized Latent Space (CLS) with the Joint Multi-Human Latent Space (JMLS) on the InterHuman dataset in terms of Reconstruction FID, MPJPE (Mean Per Joint Position Error), and MROE (Mean Relative Orientation Error), measuring reconstruction quality.
For a fair comparison, we implement a motion VAE that directly encodes two-human motion trajectories as the JMLS baseline.
The results demonstrate that CLS substantially outperforms JMLS in reconstruction quality (Recon FID: 0.307 \textit{vs.} 7.783), highlighting that canonicalization enables more effective modeling of local human motion.

As shown in Tab.~\ref{tab:ablation_studies}, we evaluate the contributions of HINT’s key components, including the canonicalized latent space (CLS), local conditions (L), and global conditions (G). 
Replacing CLS with JMLS leads to a severe degradation in generation quality, with FID increasing from 3.100 to 5.274, underscoring its necessity.
For local conditions, we remove individual components to assess their effectiveness. 
Excluding the history motion, step index, relative history, and word-level text embeddings results in slight R@Top3 drops of {0.012, 0.014, 0.025, and 0.000 (unchanged)}, respectively, compared to the full HINT (0.672). 
However, the corresponding FID values worsen significantly by {1.497, 0.124, 1.474, and 0.195}.
These consistent degradations verify that each local condition term provides complementary temporal or semantic cues and is indispensable for improving text-motion alignment and motion fidelity.
For global conditions, the exclusion of compositional command embedding decreases R@Top3 by 0.003 and worsens FID by 0.241.
Removing sequence index and total frame number has an even larger impact, with R@Top3 dropping by {0.005} and FID increasing by {0.443}. 
These results highlight that both structural sequence information and compositional commands play crucial roles in ensuring coherent long-horizon motion generation and semantically grounded interaction synthesis.

%% file: tables/results.tex
\begin{table}[t]
\caption{Results on InterHuman and InterX. $\rightarrow$ denotes closer to ground truth is better, $\uparrow/\downarrow$ means higher/lower is better, 
$\pm$ indicates the $95\%$ confidence interval. \textbf{Bold} denotes the best result.
InterMask* is the online version of InterMask, while DART$^{\dagger}$ is the two-human version of DART.}
\label{tab:quantitative_interhuman_interx_combined}
\centering
\resizebox{0.92\textwidth}{!}{%
\tablestyle{2pt}{1.01}
\begin{tabular}{@{}cllcccc@{}}
\toprule
Dataset & Setting & Method & R@Top3$\uparrow$ & FID$\downarrow$ & MM Dist$\downarrow$ & Diversity$\rightarrow$ \\
\midrule
\multirow{11}{*}{\makecell[l]{Inter\\Human}}
&  & Ground Truth    & $0.701^{\pm .008}$ & $0.273^{\pm .007}$ & $3.755^{\pm .008}$ & $7.948^{\pm .064}$ \\
\cmidrule(lr){2-7}
& \multirow{7}{*}{offline}
& T2M \citep{guo2022generating}               & $0.464^{\pm .014}$ & $13.769^{\pm .072}$ & $5.731^{\pm .013}$ & $7.046^{\pm .022}$ \\
& & MDM \citep{tevethuman}            & $0.339^{\pm .012}$ & $9.167^{\pm .056}$  & $7.125^{\pm .018}$ & $7.602^{\pm .045}$ \\
& & ComMDM \citep{shafir2024human}         & $0.466^{\pm .010}$ & $7.069^{\pm .054}$  & $6.212^{\pm .021}$ & $7.244^{\pm .038}$ \\
& & InterGen \citep{liang2024intergen}        & $0.624^{\pm .010}$ & $5.918^{\pm .079}$  & $5.108^{\pm .014}$ & $7.387^{\pm .029}$ \\
& & MoMat\textendash MoGen \citep{cai2024digital} & $0.666^{\pm .004}$ & $5.674^{\pm .085}$  & $\mathbf{3.790}^{\pm .001}$ & $8.021^{\pm .350}$ \\
& & in2IN \citep{ruiz2024in2in}           & $0.662^{\pm .009}$ & $5.535^{\pm .120}$  & $3.803^{\pm .002}$ & $7.953^{\pm .047}$ \\
& & \textbf{InterMask} \citep{javedintermask} & $\mathbf{0.683}^{\pm .004}$ & $\mathbf{5.154}^{\pm .061}$ & $\mathbf{3.790}^{\pm .002}$ & $\mathbf{7.944}^{\pm .033}$ \\
\cmidrule(lr){2-7}
& \multirow{3}{*}{online}
& InterMask*        & $0.557^{\pm .004}$ & $14.352^{\pm .133}$ & $3.852^{\pm .001}$ & $7.485^{\pm .032}$ \\
& & DART$^{\dagger}$& $0.642^{\pm .005}$ & $4.979^{\pm .053}$  & $3.813^{\pm .001}$ & $7.950^{\pm .032}$ \\
& & \textbf{HINT}  & $\mathbf{0.672}^{\pm .004}$ & $\mathbf{3.100}^{\pm .035}$  & $\mathbf{3.796}^{\pm .001}$ & $\mathbf{7.898}^{\pm .023}$ \\
\midrule
\midrule
\multirow{9}{*}{InterX}
&  & Ground Truth    & $0.736^{\pm .003}$ & $0.002^{\pm .0002}$ & $3.536^{\pm .013}$ & $9.734^{\pm .078}$ \\
\cmidrule(lr){2-7}
& \multirow{5}{*}{offline}
& T2M \citep{guo2022generating}               & $0.396^{\pm .005}$ & $5.481^{\pm .382}$  & $9.576^{\pm .006}$ & $2.771^{\pm .151}$ \\
& & MDM \citep{tevethuman}             & $0.426^{\pm .005}$ & $23.701^{\pm .057}$ & $9.548^{\pm .014}$ & $5.856^{\pm .077}$ \\
& & ComMDM \citep{shafir2024human}          & $0.236^{\pm .004}$ & $29.266^{\pm .067}$ & $6.870^{\pm .017}$ & $4.734^{\pm .067}$ \\
& & InterGen \citep{liang2024intergen}        & $0.429^{\pm .005}$ & $5.207^{\pm .216}$  & $9.580^{\pm .011}$ & $7.788^{\pm .208}$ \\
& & \textbf{InterMask} \citep{javedintermask} & $\mathbf{0.705}^{\pm .005}$ & $\mathbf{0.399}^{\pm .013}$ & $\mathbf{3.705}^{\pm .017}$ & $\mathbf{9.046}^{\pm .073}$ \\
\cmidrule(lr){2-7}
& \multirow{3}{*}{online}
& InterMask*        & $0.169^{\pm .003}$ & $19.445^{\pm .199}$ & $7.885^{\pm .003}$ & $6.250^{\pm .007}$ \\
& & DART$^{\dagger}$& $0.510^{\pm .003}$ & $8.600^{\pm .075}$  & $5.492^{\pm .014}$ & $8.405^{\pm .073}$ \\
& & \textbf{HINT}   & $\mathbf{0.682}^{\pm .003}$ & $\mathbf{0.278}^{\pm .012}$ & $\mathbf{4.007}^{\pm .016}$ & $\mathbf{8.886}^{\pm .066}$ \\
\bottomrule
\end{tabular}}
\end{table}

%% file: tables/abla_ih_key_components.tex
\begin{table}[t]
\centering
\caption{Ablations HINT's key components on InterHuman.
\textbf{L}/\textbf{G} indicates local/global conditions.
% \textbf{Bold} means the best results.
}
\label{tab:ablation_studies}
% \resizebox{\textwidth}{!}
{%
\tablestyle{3pt}{1.01}
\begin{tabular}{@{}clcccc@{}}
\toprule
 & \multirow{1}{*}{Method} & R@Top3$\uparrow$ & FID$\downarrow$ & MM Dist$\downarrow$ & Diversity$\rightarrow$ \\
\midrule
\multicolumn{2}{l}{Ground Truth}  & $0.701^{\pm .008}$ & $0.273^{\pm .007}$ & $3.755^{\pm .008}$ & $7.948^{\pm .064}$ \\
\midrule
\multicolumn{2}{l}{w/o Canonicalized Latent Space}  & ${0.633}^{\pm .006}$ & ${5.274}^{\pm .051}$  & ${3.814}^{\pm .001}$ & ${7.802}^{\pm .025}$ \\
\midrule
\multirow{4}{*}{\textbf{L}}
& w/o History Motion Embedding   & ${0.660}^{\pm .005}$ & ${4.597}^{\pm .063}$  & ${3.802}^{\pm .001}$ & ${7.849}^{\pm .030}$ \\
 & w/o Step Index Embedding       & ${0.658}^{\pm .005}$ & ${3.224}^{\pm .044}$  & ${3.802}^{\pm .001}$ & ${7.875}^{\pm .030}$ \\
 & w/o Relative History Embedding & ${0.647}^{\pm .006}$ & ${4.574}^{\pm .058}$  & ${3.808}^{\pm .001}$ & $\mathbf{7.912}^{\pm .031}$ \\
 & w/o Word-level Text Embedding & $\mathbf{0.672}^{\pm .004}$ & ${3.295}^{\pm .049}$  & ${3.798}^{\pm .001}$ & ${7.908}^{\pm .031}$ \\
\midrule
\multirow{2}{*}{\textbf{G}}
 & \makecell[l]{w/o Sequence Index \\ ~~~~~~~\& Total Frame Number} & ${0.667}^{\pm .004}$ & ${3.543}^{\pm .058}$  & ${3.800}^{\pm .001}$ & ${7.874}^{\pm .024}$ \\
 & w/o Compositional Command Embedding & ${0.669}^{\pm .003}$ & ${3.341}^{\pm .045}$  & ${3.797}^{\pm .001}$ & ${7.879}^{\pm .024}$ \\
\midrule
\multicolumn{2}{l}{\textbf{HINT}} & $\mathbf{0.672}^{\pm .004}$ & $\mathbf{3.100}^{\pm .035}$  & $\mathbf{3.796}^{\pm .001}$ & ${7.898}^{\pm .023}$ \\
\bottomrule
\end{tabular}}
\end{table}

%% file: tables/abla_ih_cls.tex
\begin{wraptable}{r}{0.45\textwidth}
  \centering
  \vspace{-0.4cm}
  \caption{Ablation of the Canonicalized Latent Space on InterHuman.}
  \label{tab:ablation_studies_MVAE}
  \tablestyle{2pt}{1.01}
  \resizebox{0.45\textwidth}{!}{%
    \begin{tabular}{@{}lccc@{}}
      \toprule
      Method & Recon FID$\downarrow$ & MPJPE$\downarrow$ & MROE$\downarrow$ \\
      \midrule
      {Joint Multi-Human Latent Space} & $7.783^{\pm .006}$ & $0.213^{\pm .001}$ & $0.426^{\pm .001}$ \\
      \textbf{{Canonicalized Latent Space}} & $\mathbf{0.307}^{\pm .005}$ & $\mathbf{0.138}^{\pm .001}$ & $\mathbf{0.118}^{\pm .002}$ \\
      \bottomrule
      \vspace{-0.7cm}
    \end{tabular}
  }
\end{wraptable}

%% file: sections/9_conclusion.tex
\section{Conclusion}
In this paper, we presented HINT, the first autoregressive framework for multi-human motion generation with hierarchical interaction modeling in diffusion. 
By disentangling local motion semantics from inter-person interactions in a canonicalized latent space and adopting a sliding-window strategy that integrates both local and global context, HINT effectively adapts to varying numbers of human participants while maintaining long-horizon coherence. 
Extensive experiments on public benchmarks demonstrate that HINT not only matches the performance of strong offline models but also significantly outperforms existing autoregressive baselines.
In the future, an exciting direction is to extend our framework to incorporate objects and environments, enabling multi-human motion generation with object interactions. 
Text-driven generation of complex multi-agent behaviors in dynamic scenes remains a highly challenging yet impactful problem, and we believe HINT provides a strong foundation for advancing this line of research.



%% file: sections/10_appendix.tex
\clearpage

\renewcommand\thefigure{\thesection-\arabic{figure}}
\renewcommand\thetable{\thesection-\arabic{table}}
\renewcommand\theequation{\thesection-\arabic{equation}}
\setcounter{figure}{0} 
\setcounter{table}{0}
\setcounter{equation}{0}

\appendix
\section{Extension to Multi-Human Motion Generation}
\begin{figure}[h]
    \centering    
    \includegraphics[width=\linewidth]{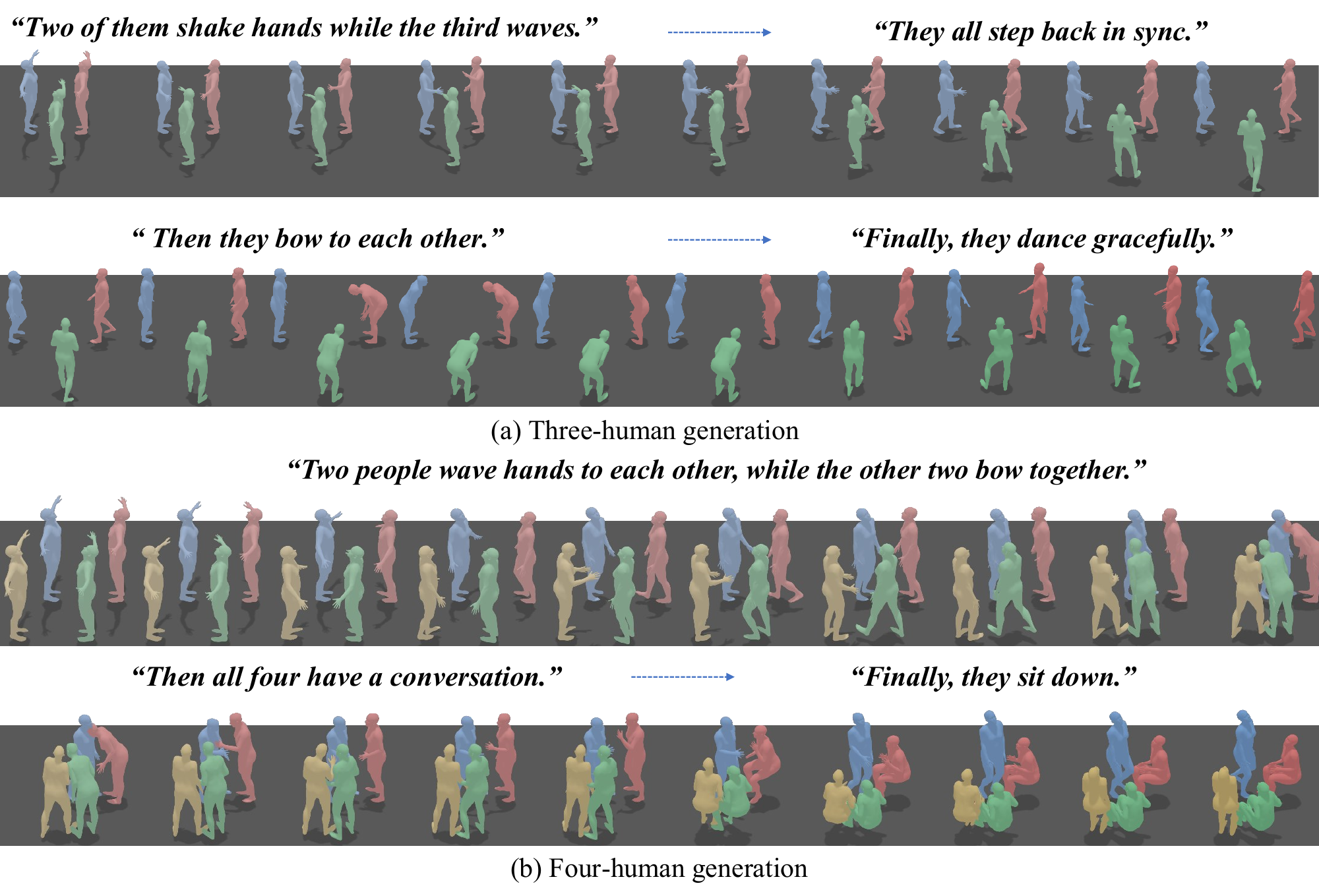}
    \captionsetup{justification=centering}
    \caption{Additional examples of three-human motion generation result.}
    \label{fig:add_ma}
\end{figure}

Our method can be naturally extended to multi-human interaction scenarios. Building upon the two-person generation framework, we simply incorporate the motion histories of additional participants into the conditioning to achieve joint modeling of multiple agents. 
Since large-scale, high-quality multi-human interaction datasets are currently lacking, we directly apply this extension based on the model trained on two-person data. 
Fig.~\ref{fig:add_ma} presents an additional example of three-human motion generation results.
For further improvements in generation quality and interaction details, the Interaction Cross-Attention module can be lightly fine-tuned once more comprehensive multi-human datasets become available.

\section{Implementation Details}
We provide more details of the model architecture, training, and the compared baselines. The implementation code of HINT is provided in the supplementary material as an attachment.
\subsection{Model Architecture}
\textbf{Motion VAE.} The Motion VAE adopts a transformer-based encoder–decoder architecture. 
Both encoder and decoder are constructed from stacked Transformer layers with residual connections and learned positional encodings. 
Raw motion sequences (history and future) are first linearly projected into a hidden space, and concatenated with a set of learnable global motion tokens. 
The encoder outputs the mean and variance of a Gaussian distribution, from which latent variables are sampled using the reparameterization trick. 
For the decoder, two variants are supported: we use all-encoder structure, where latent vectors and history embeddings are concatenated with query tokens and passed through a symmetric Transformer encoder. 

\textbf{Interaction-Aware Diffusion.}The denoiser consists of $L_{diff}=8$ transformer blocks with $H=4$ heads, hidden size $d=512$, and feed-forward width $d_{\mathrm{ff}}=1024$.
It incorporates our Hierarchical Motion Condition (HMC), which fuses \textbf{local} conditions (individual history, step index, token-level text embedding, partners' history), \textbf{global} conditions (sequence length, sentence-level text embedding), through self- and cross-attention, enabling both fine-grained alignment and global consistency.

Tab. \ref{tab:motion_vae_params} shows the details model parameters of 
% MVAE
Motion VAE
and Interaction-Aware Diffusion.
\input{tables/MVAE_params}

\subsection{Training Details}
\label{App:training details}
\textbf{Three-stage {Training Strategy} for {Motion VAE}.}
We adopt a three-stage training strategy for the 
% MVAE
{Motion VAE}
and Interaction-Aware Diffusion following DART \citep{zhaodartcontrol}:

{\textit{Stage I (Ground-Truth History)}}: The model is trained on motion windows fully extracted from ground-truth sequences.

{\textit{Stage II (Mixed History)}}: We gradually introduce predicted windows as part of the historical context. Specifically, the probability of replacing ground-truth history with the model’s predictions is linearly increased during training.

{\textit{Stage III (Predicted History)}}: The model is trained with history composed entirely of its own predicted windows, ensuring robustness in fully autoregressive generation.

\textbf{Detailed Loss Definition for Interaction-Aware Diffusion.}
Following InterGen \citep{liang2024intergen}, the detailed {definition} of regularization loss {is} as follows:

\textit{1) Joint affinity}  
\begin{equation}
    \mathcal{L}_{\text{aff}} =
\left\| 
\Big( D(m_A, m_B) - D(\hat{m}_A, \hat{m}_B) \Big) 
\odot \mathbf{I}\!\left( D(m_A, m_B) < \overline{D}_1 \right) 
\right\|_2^2,
\end{equation}
where $D(\cdot)$ computes the pairwise joint distance matrix, $\mathbf{I}(\cdot)$ is the indicator function, and $\overline{D}_1$ denotes a predefined distance threshold.
This loss encourages the predicted motion $(\hat{m}_A, \hat{m}_B)$ to preserve the joint-level spatial affinity observed in the ground truth $(m_A, m_B)$.

\textit{2) Distance map}  
\begin{equation}
    \mathcal{L}_{\text{dist}} =
    \left\| 
    \big(D(m_A, m_B) - D(\hat{m}_A, \hat{m}_B)\big) 
    \odot \mathbf{I} \big( D(\hat{m}_A, \hat{m}_B)  < \overline{D}_2 \big) 
    \right\|_2^2,
\end{equation}
$\mathcal{L}_{\text{dist}}$ enforces accurate modeling of close-range spatial relationships while ignoring distant pairs that are less critical for interaction.

\textit{3) Relative orientation}  
\begin{equation}
    \mathcal{L}_{\text{ori}} = 
    \left\| 
    O\!\left(m_A, m_B\right) - 
    O\!\left(\hat{m}_A, \hat{m}_B\right) 
    \right\|_2^2,
\end{equation}
where $O(\cdot_A,\cdot_B)$ denotes the 6D representation \citep{rot6d_zhou2019} of the relative rotation matrix from human A to B. 
$\mathcal{L}_{\text{ori}}$ enforces the predicted motions to preserve the relative orientations between the two humans, ensuring coherent and physically plausible interactions.

We also use a truncated regularization strategy: the regularization loss is only applied at lower diffusion timesteps. 
This prevents the denoiser from being biased towards implausible averaged poses and ensures more realistic motion generation.

Tab. \ref{tab:hyperparams} presents the key hyperparameters of Motion VAE and Interaction-Aware Diffusion.

\subsection{Details of Compared Methods} \label{appendix:baseline_details}
The details of baselines are as follows: 

\textbf{Offline Methods.} 
T2M \citep{guo2022generating} is a Transformer-based motion generation framework that formulates text-conditioned motion synthesis as a sequence-to-sequence problem in a learned motion latent space. 
MDM \citep{tevethuman} employs a diffusion-based approach that conditions the denoising process on text or other modalities to generate high-quality, temporally coherent motions. 
ComMDM \citep{shafir2024human} adds a lightweight communication block between two frozen MDM pretrained models to enable few-shot human-human interaction generation. 
InterGen \citep{liang2024intergen} incorporates a mutual attention mechanism into the diffusion process to explicitly model inter-person dependencies for multi-person interaction generation. 
MoMat-MoGen \citep{cai2024digital} combines motion matching with generative modeling to enhance diversity while preserving motion naturalness, enabling high-quality text-driven motion synthesis. 
in2IN \citep{ruiz2024in2in} leverages both individual motion descriptions and global interaction semantics to improve diversity and accuracy in human–human interaction generation. 
InterMask \citep{javedintermask} encodes motions as 2D token maps and jointly predicts masked tokens for both characters, enabling high-fidelity and diverse interaction generation.
We use the results reported in their original papers.

\textbf{Online Extensions.} For InterMask*, we retain the 2D VQ-VAE structure of the original InterMask \citep{javedintermask}, but retrain it under our sliding-window setting to adapt to online generation. 
On top of this representation, we employ our autoregressive framework: within each prediction window, InterMask is used for motion generation. During generation, the history morion remains unmasked.
And for DART$^{\dagger}$, we retrain DART \citep{zhaodartcontrol} on the InterHuman \citep{liang2024intergen} and InterX \citep{xu2024inter} dataset, where both humans share the same network during generation.

\section{Additional Visualization Results} \label{addition results}
\input{tables/detailed_res}
\input{tables/detailed_abla}
\textbf{Quantitative Results.} Detailed R-Precision results for InterHuman and InterX are presented in Tab.~\ref{tab:quantitative_interhuman_interx_rprec_only}.
The ablation R-Precision results are reported in Tab.~\ref{tab:R_precision_ablation_studies}.

\begin{figure}[h]
    \centering    \includegraphics[width=\linewidth]{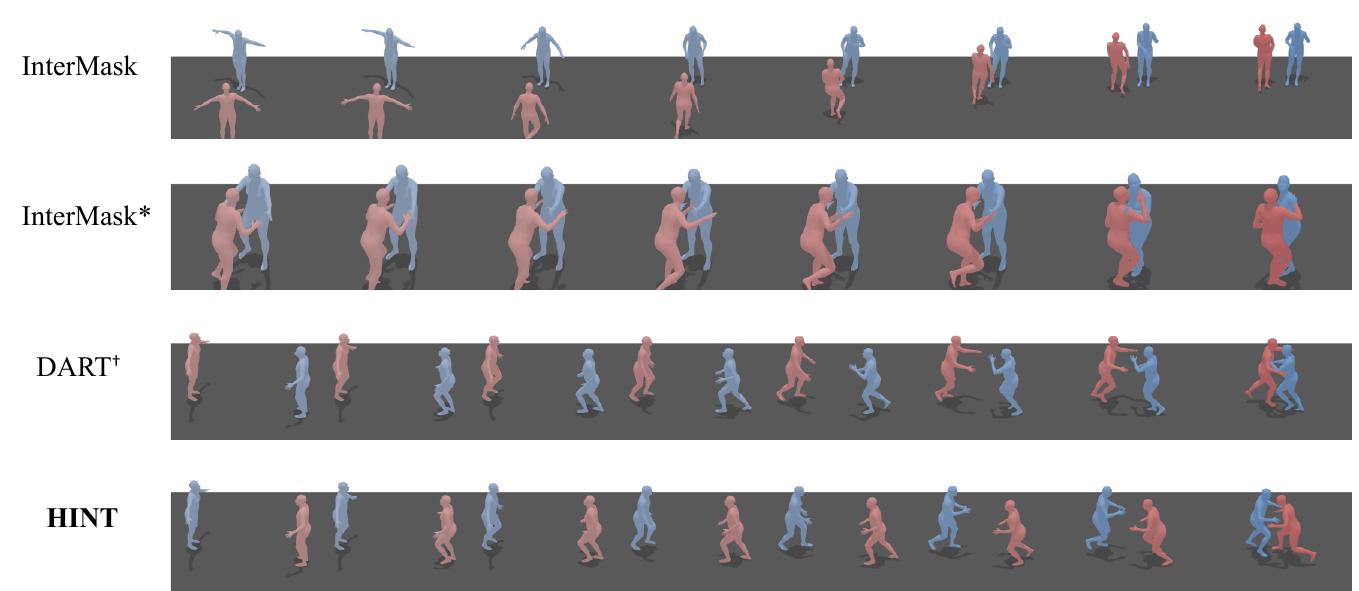}
    \captionsetup{justification=centering}
    \caption{They rush towards each other.}
    \label{fig:add_1}
\end{figure}

\begin{figure}[h]
    \centering    \includegraphics[width=\linewidth]{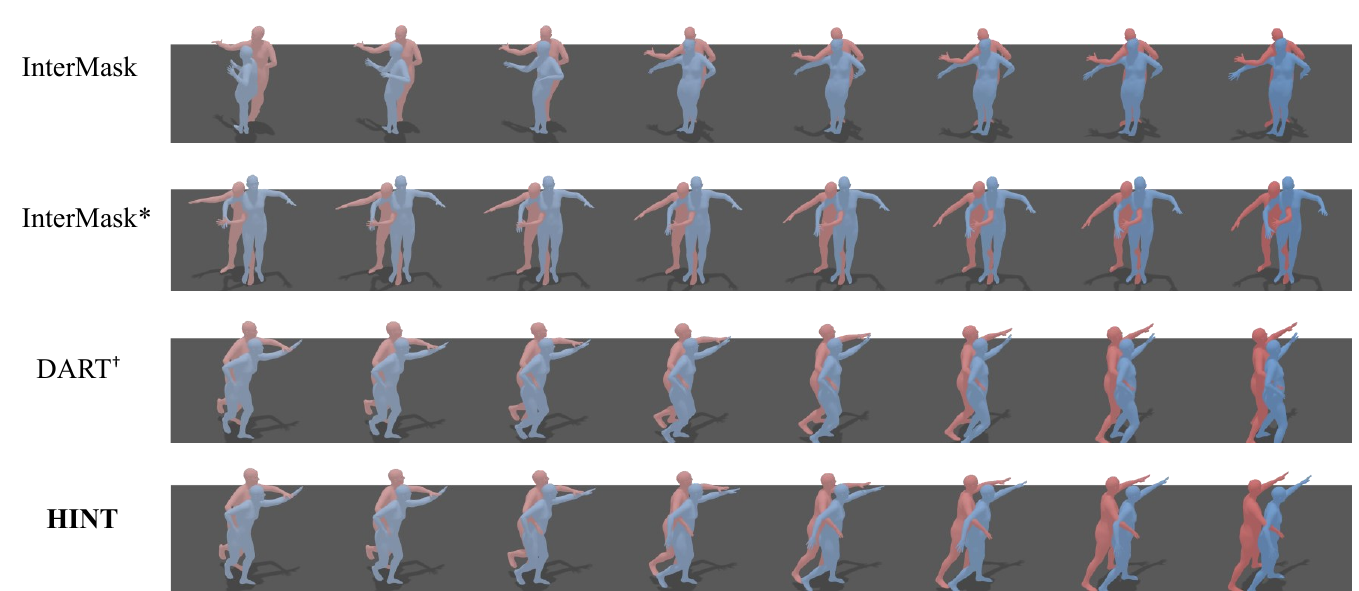}
    \captionsetup{justification=centering}
    \caption{The two people take a small step to the right side with their right foot.}
    \label{fig:add_2}
\end{figure}

\begin{figure}[h]
    \centering    \includegraphics[width=\linewidth]{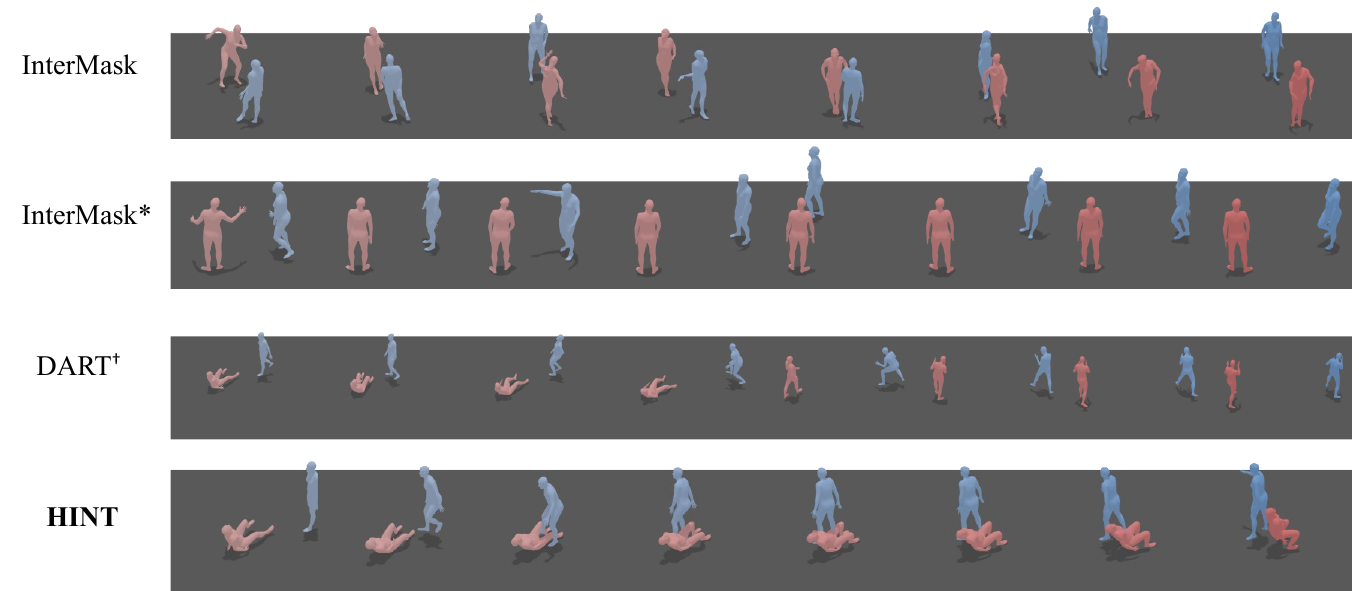}
    \captionsetup{justification=centering}
    \caption{The first one walks towards the second one and takes a few steps, pretends to hit the second one with the right hand, and then walks away to the side.}
    \label{fig:add_3}
\end{figure}

\begin{figure}[h]
    \centering    \includegraphics[width=\linewidth]{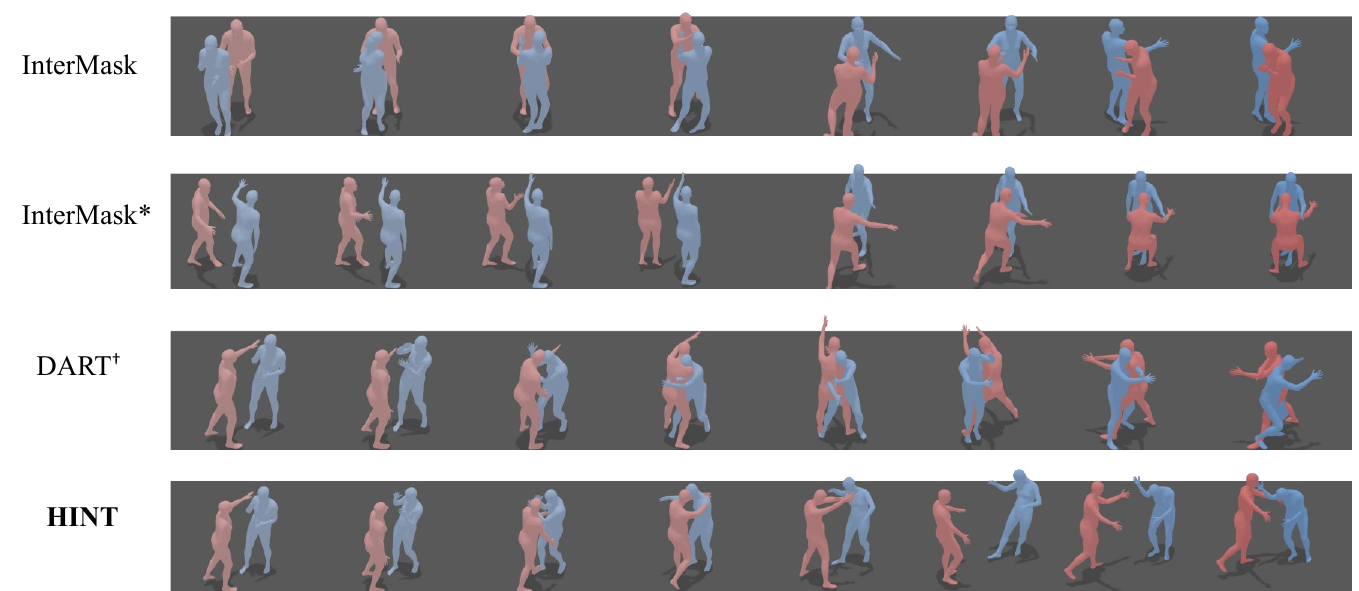}
    \captionsetup{justification=centering}
    \caption{One person extends their left arm and pushes the other person's right arm, and then they push each other back and forth.}
    \label{fig:add_4}
\end{figure}

\begin{figure}[h]
    \centering    \includegraphics[width=\linewidth]{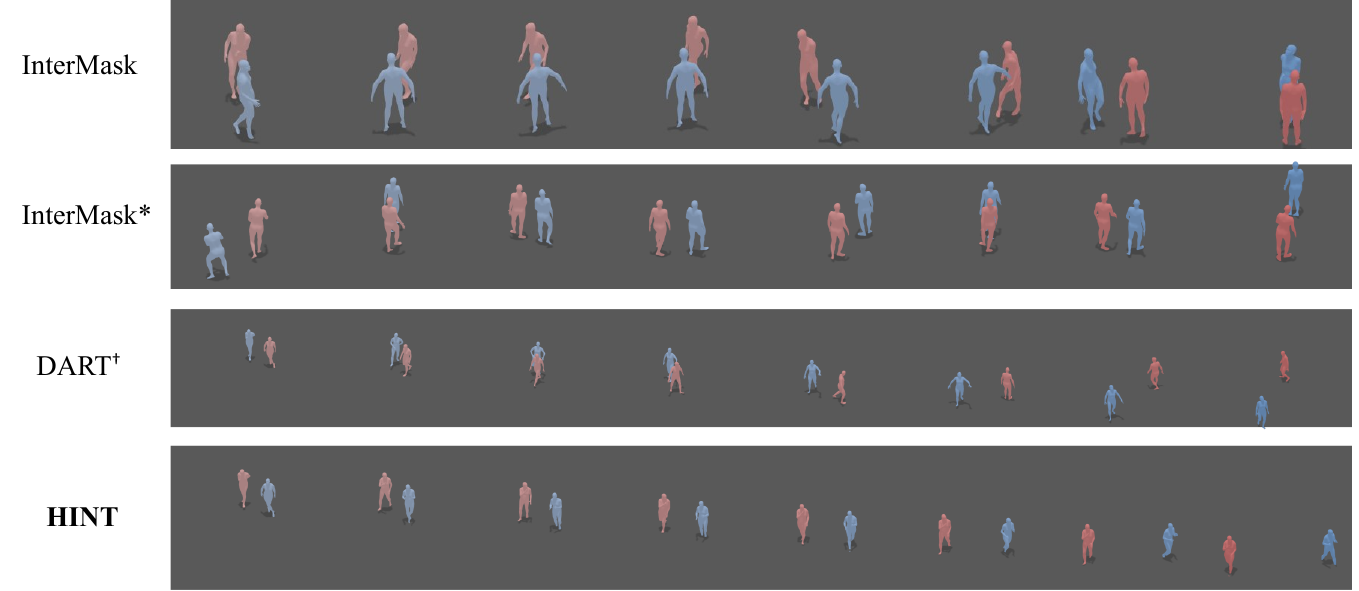}
    \captionsetup{justification=centering}
    \caption{Two people proceed ahead together.}
    \label{fig:add_5}
\end{figure}

\begin{figure}[h]
    \centering    \includegraphics[width=\linewidth]{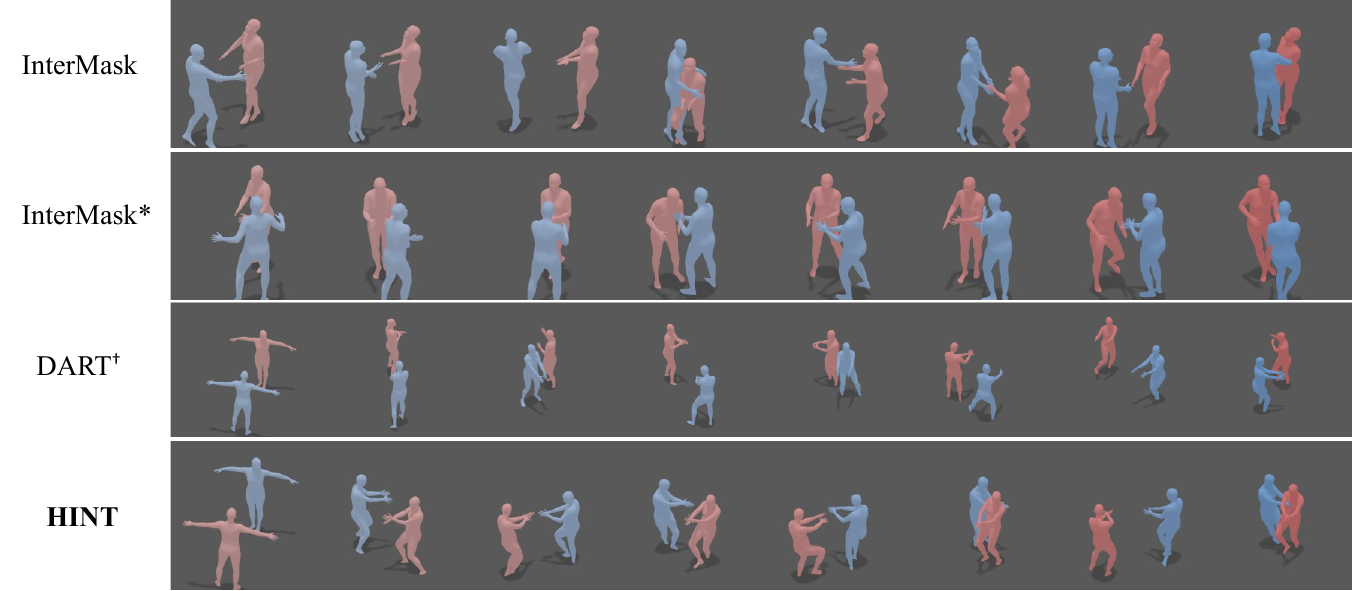}
    \captionsetup{justification=centering}
    \caption{Both they encircle joining hands.}
    \label{fig:add_6}
\end{figure}

\textbf{Qualitative Results.}
More visualization results are shown in Figs.~\ref{fig:add_1}--\ref{fig:add_6}.

\section{{Details of} User Study}
\begin{figure}[h]
    \centering    
    \includegraphics[width=\linewidth]{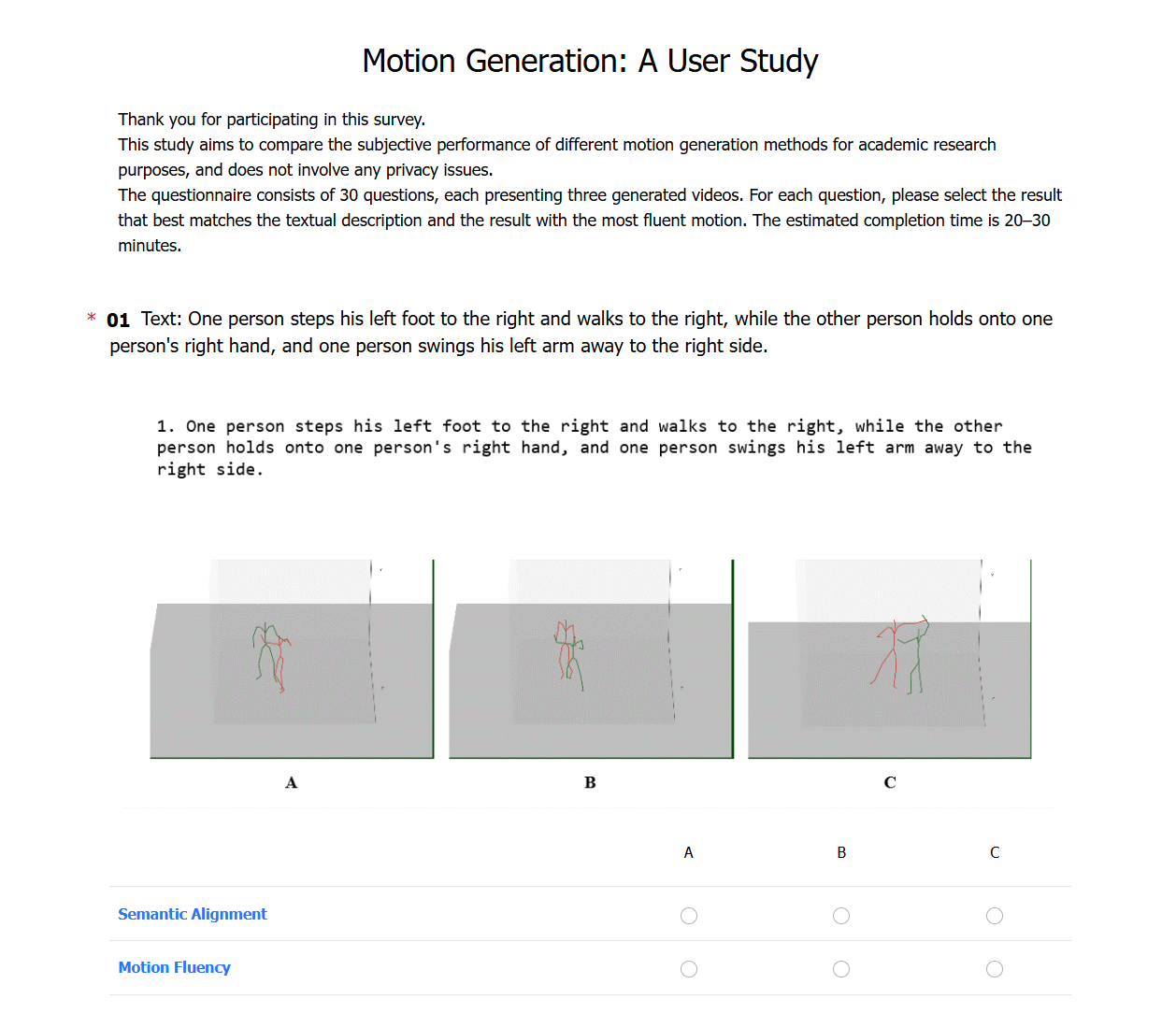}
    \captionsetup{justification=centering}
    \caption{Interface of the User Study.}
    \label{fig:questionnarie}
\end{figure}

We randomly sampled 30 textual descriptions from the test set of InterHuman~\citep{liang2024intergen}. 
For each description, motion videos of the same length as the ground truth are generated using HINT, DART$^\dagger$, and InterMask~\citep{javedintermask}.
An online questionnaire is then distributed, where participants viewed the text and the corresponding videos and selected the best video based on semantic alignment and motion
fluency. 
In total, 15 participants completed the survey.
Fig.~\ref{fig:questionnarie} shows a screenshot of the questionnaire interface.

\section{Use of LLMs}
This paper used large language models (LLMs) to assist with language polishing. 
No core ideas, analyses, or experimental results were generated by LLMs.

%% file: tables/MVAE_params.tex
\begin{table}[!htbp]
\begin{minipage}{0.48\linewidth}
\centering
\caption{Parameters of Motion VAE and Interaction-Aware Diffusion.}
\label{tab:motion_vae_params}
\tablestyle{13pt}{1.01}
\begin{tabular}{@{}lc@{}}
\toprule
\textbf{Parameter} & \textbf{Value} \\
\midrule
Latent dim ($d_z$)         & 256 \\
Hidden dim ($d_h$)         & 512 \\
Feed-forward dim ($d_{\mathrm{ff}}$)& 1024 \\
Layers for VAE ($L_{VAE}$)               & 5 \\
\makecell[l]{Transformer blocks \\ of diffusion ($L_{diff}$)}               & 8 \\
Attention heads ($H$)      & 4 \\
Dropout ($p$)              & 0.1 \\
CLIP version               & ViT-L/14@336px \\
\bottomrule
\end{tabular}
\end{minipage}\hfill
\begin{minipage}{0.48\linewidth}
\centering
\caption{Training hyperparameters for Motion VAE and Interaction-Aware Diffusion.}
\label{tab:hyperparams}
\tablestyle{40pt}{1.01}
\begin{tabular}{@{}lc@{}}
\toprule
\textbf{Hyperparameter} & \textbf{Value} \\
\midrule
stage1\_steps       & 100,000 \\
stage2\_steps       & 100,000 \\
stage3\_steps       & 100,000 \\
learning\_rate      & $10^{-4}$    \\
$\beta$             & $10^{-4}$    \\
$\lambda_\text{aff}$     & $10^{-1}$    \\
$\lambda_\text{dist}$    & $10^{-1}$    \\
$\lambda_\text{ori}$     & $10^{-4}$    \\
$\overline{D}_1$     & $10^{-1}$    \\
$\overline{D}_2$     & 1.0    \\
\bottomrule
\end{tabular}
\end{minipage}
\end{table}

%% file: tables/detailed_res.tex
\begin{table}[t]
\caption{Detailed R-precision results on InterHuman and InterX. 
\textbf{Bold} denotes the best result for each setting.}
\label{tab:quantitative_interhuman_interx_rprec_only}
\centering
\resizebox{\textwidth}{!}{%
\tablestyle{15pt}{1.01}
\begin{tabular}{@{}cllccc@{}}
\toprule
 \multirow{2}{*}{Dataset} & \multirow{2}{*}{Setting} & \multirow{2}{*}{Method} & \multicolumn{3}{c}{R-Precision$\uparrow$} \\
\cmidrule(lr){4-6}
& & & Top 1 & Top 2 & Top 3 \\
\midrule
\multirow{11}{*}{\makecell[l]{Inter\\Human}}
&  & Ground Truth & $0.452^{\pm .008}$ & $0.610^{\pm .009}$ & $0.701^{\pm .008}$ \\
\cmidrule(lr){2-6}
& \multirow{7}{*}{offline}
& T2M \citep{guo2022generating}            & $0.238^{\pm .012}$ & $0.325^{\pm .010}$ & $0.464^{\pm .014}$ \\
& & MDM \citep{tevethuman}          & $0.153^{\pm .012}$ & $0.260^{\pm .009}$ & $0.339^{\pm .012}$ \\
& & ComMDM \citep{shafir2024human}       & $0.223^{\pm .009}$ & $0.334^{\pm .008}$ & $0.466^{\pm .010}$ \\
& & InterGen \citep{liang2024intergen}     & $0.371^{\pm .010}$ & $0.515^{\pm .012}$ & $0.624^{\pm .010}$ \\
& & MoMat\textendash MoGen \citep{cai2024digital} & $\mathbf{0.449}^{\pm .004}$ & $0.591^{\pm .003}$ & $0.666^{\pm .004}$ \\
& & in2IN \citep{ruiz2024in2in}        & $0.425^{\pm .008}$ & $0.576^{\pm .008}$ & $0.662^{\pm .009}$ \\
& & \textbf{InterMask} \citep{javedintermask}     & $\mathbf{0.449}^{\pm .004}$ & $\mathbf{0.599}^{\pm .005}$ & $\mathbf{0.683}^{\pm .004}$ \\
\cmidrule(lr){2-6}
& \multirow{3}{*}{online}
& InterMask*     & $0.331^{\pm .005}$ & $0.471^{\pm .005}$ & $0.557^{\pm .004}$ \\
& & DART$^{\dagger}$     & $0.395^{\pm .005}$ & $0.553^{\pm .005}$ & $0.642^{\pm .005}$ \\
& & \textbf{HINT} & $\mathbf{0.432}^{\pm .004}$ & $\mathbf{0.587}^{\pm .004}$ & $\mathbf{0.672}^{\pm .004}$ \\
\midrule
\midrule
\multirow{9}{*}{InterX}
&  & Ground Truth & $0.429^{\pm .004}$ & $0.626^{\pm .003}$ & $0.736^{\pm .003}$ \\
\cmidrule(lr){2-6}
& \multirow{5}{*}{offline}
& T2M \citep{guo2022generating}           & $0.184^{\pm .010}$ & $0.298^{\pm .006}$ & $0.396^{\pm .005}$ \\
& & MDM \citep{tevethuman}          & $0.203^{\pm .009}$ & $0.329^{\pm .007}$ & $0.426^{\pm .005}$ \\
& & ComMDM \citep{shafir2024human}       & $0.090^{\pm .002}$ & $0.165^{\pm .004}$ & $0.236^{\pm .004}$ \\
& & InterGen \citep{liang2024intergen}     & $0.207^{\pm .004}$ & $0.335^{\pm .005}$ & $0.429^{\pm .005}$ \\
& & \textbf{InterMask} \citep{javedintermask}   & $\mathbf{0.403}^{\pm .005}$ & $\mathbf{0.595}^{\pm .004}$ & $\mathbf{0.705}^{\pm .005}$ \\
\cmidrule(lr){2-6}
& \multirow{3}{*}{online}
& InterMask*     & $0.061^{\pm .004}$ & $0.119^{\pm .003}$ & $0.169^{\pm .003}$ \\
& & DART$^{\dagger}$     & $0.252^{\pm .003}$ & $0.402^{\pm .003}$ & $0.510^{\pm .003}$ \\
& & \textbf{HINT} & $\mathbf{0.386}^{\pm .005}$ & $\mathbf{0.572}^{\pm .004}$ & $\mathbf{0.682}^{\pm .003}$ \\
\bottomrule
\end{tabular}}
\end{table}

%% file: tables/detailed_abla.tex
\begin{table}[t]
\centering
\caption{Detailed R-precision results of ablation studies on InterHuman.
\textbf{L} and \textbf{G} indicate local and global conditions, respectively.
% \textbf{Bold} denotes the best results.
}
\label{tab:R_precision_ablation_studies}
{%
\tablestyle{3pt}{1.01}
\begin{tabular}{@{}clccc@{}}
\toprule
\multirow{2}{*}{ } & \multirow{2}{*}{Method} & \multicolumn{3}{c}{R Precision$\uparrow$} \\
\cmidrule(lr){3-5}
 & & R@Top1 & R@Top2 & R@Top3 \\
\midrule
\multicolumn{2}{l}{Ground Truth}  & $0.452^{\pm .008}$ & $0.610^{\pm .009}$ & $0.701^{\pm .008}$ \\
\midrule
\multicolumn{2}{l}{w/o Canonicalized Latent Space}  & ${0.396}^{\pm .005}$ & ${0.548}^{\pm .005}$ & ${0.633}^{\pm .006}$ \\
\midrule
\multirow{4}{*}{\textbf{L}} 
 & w/o History Motion Embedding   & ${0.421}^{\pm .006}$ & ${0.576}^{\pm .005}$ & ${0.660}^{\pm .005}$ \\
 & w/o Step Index Embedding       & ${0.413}^{\pm .005}$ & ${0.570}^{\pm .007}$ & ${0.658}^{\pm .005}$ \\
 & w/o Relative History Embedding  & ${0.405}^{\pm .005}$ & ${0.563}^{\pm .005}$ & ${0.647}^{\pm .006}$ \\
 & w/o Word-level Text Embedding & ${0.429}^{\pm .006}$ & $\mathbf{0.591}^{\pm .006}$ & $\mathbf{0.672}^{\pm .004}$ \\
\midrule
\multirow{2}{*}{\textbf{G}} 
 & \makecell[l]{w/o Sequence Index \\ \& Total Frames Embedding} & ${0.425}^{\pm .004}$ & ${0.584}^{\pm .004}$ & ${0.667}^{\pm .004}$ \\
 & w/o Compositional Command Embedding & ${0.420}^{\pm .005}$ & ${0.582}^{\pm .003}$ & ${0.669}^{\pm .003}$ \\
\midrule
\multicolumn{2}{l}{\textbf{HINT}} & $\mathbf{0.432}^{\pm .004}$ & ${0.587}^{\pm .004}$ & $\mathbf{0.672}^{\pm .004}$ \\
\bottomrule
\end{tabular}}
\end{table}

%% file: ref.bib
@string{TOG="ACM Transactions on Graphics (TOG)"}

@STRING{CVPR="IEEE/CVF Conference on Computer Vision and Pattern Recognition (CVPR)"}

@string{ICCV="IEEE/CVF International Conference on Computer Vision (ICCV)"}

@STRING{ECCV="European Conference on Computer Vision (ECCV)"}

@string{ICLR="International Conference on Learning Representations (ICLR)"}

@string{NIPS="Conference on Neural Information Processing Systems (NeurIPS)"}

@string{IJCV="International Journal of Computer Vision (IJCV)"}

@inproceedings{petrovich2022temos,
  title={Temos: Generating diverse human motions from textual descriptions},
  author={Petrovich, Mathis and Black, Michael J and Varol, G{\"u}l},
  booktitle=ECCV,
  pages={480--497},
  year={2022},
  organization={Springer}
}

@inproceedings{chen2024taming,
  title={Taming diffusion probabilistic models for character control},
  author={Chen, Rui and Shi, Mingyi and Huang, Shaoli and Tan, Ping and Komura, Taku and Chen, Xuelin},
  booktitle={ACM SIGGRAPH 2024 Conference Papers},
  pages={1--10},
  year={2024}
}

@inproceedings{peebles2023scalable,
  title={Scalable diffusion models with transformers},
  author={Peebles, William and Xie, Saining},
  booktitle=ICCV,
  pages={4195--4205},
  year={2023}
}

@article{sahili2025text,
  title={Text-driven Motion Generation: Overview, Challenges and Directions},
  author={Sahili, Ali Rida and Neji, Najett and Tabia, Hedi},
  journal={arXiv preprint arXiv:2505.09379},
  year={2025}
}

@inproceedings{guo2022generating,
  title={Generating diverse and natural 3d human motions from text},
  author={Guo, Chuan and Zou, Shihao and Zuo, Xinxin and Wang, Sen and Ji, Wei and Li, Xingyu and Cheng, Li},
  booktitle=CVPR,
  pages={5152--5161},
  year={2022}
}

@inproceedings{tevethuman,
  title={Human Motion Diffusion Model},
  author={Tevet, Guy and Raab, Sigal and Gordon, Brian and Shafir, Yoni and Cohen-or, Daniel and Bermano, Amit Haim},
  booktitle=ICLR,
  year={2022}
}

@article{zhang2024motiondiffuse,
  title={Motiondiffuse: Text-driven human motion generation with diffusion model},
  author={Zhang, Mingyuan and Cai, Zhongang and Pan, Liang and Hong, Fangzhou and Guo, Xinying and Yang, Lei and Liu, Ziwei},
  journal={IEEE transactions on pattern analysis and machine intelligence},
  volume={46},
  number={6},
  pages={4115--4128},
  year={2024},
  publisher={IEEE}
}

@article{jiang2023motiongpt,
  title={Motiongpt: Human motion as a foreign language},
  author={Jiang, Biao and Chen, Xin and Liu, Wen and Yu, Jingyi and Yu, Gang and Chen, Tao},
  journal=NIPS,
  volume={36},
  pages={20067--20079},
  year={2023}
}

@inproceedings{zhang2023generating,
    title={T2M-GPT: Generating Human Motion from Textual Descriptions with Discrete Representations},
    author={Zhang, Jianrong and Zhang, Yangsong and Cun, Xiaodong and Huang, Shaoli and Zhang, Yong and Zhao, Hongwei and Lu, Hongtao and Shen, Xi},
    booktitle=CVPR,
    year={2023},
}

@inproceedings{chen2023executing,
  title={Executing your commands via motion diffusion in latent space},
  author={Chen, Xin and Jiang, Biao and Liu, Wen and Huang, Zilong and Fu, Bin and Chen, Tao and Yu, Gang},
  booktitle=CVPR,
  pages={18000--18010},
  year={2023}
}

@inproceedings{barquero2024seamless,
  title={Seamless human motion composition with blended positional encodings},
  author={Barquero, German and Escalera, Sergio and Palmero, Cristina},
  booktitle=CVPR,
  pages={457--469},
  year={2024}
}

@inproceedings{xu2023interdiff,
  title={Interdiff: Generating 3d human-object interactions with physics-informed diffusion},
  author={Xu, Sirui and Li, Zhengyuan and Wang, Yu-Xiong and Gui, Liang-Yan},
  booktitle=ICCV,
  pages={14928--14940},
  year={2023}
}

@article{alexanderson2023listen,
  title={Listen, denoise, action! audio-driven motion synthesis with diffusion models},
  author={Alexanderson, Simon and Nagy, Rajmund and Beskow, Jonas and Henter, Gustav Eje},
  journal=TOG,
  volume={42},
  number={4},
  pages={1--20},
  year={2023},
  publisher={ACM New York, NY, USA}
}

@inproceedings{zhaodartcontrol,
  title={DartControl: A Diffusion-Based Autoregressive Motion Model for Real-Time Text-Driven Motion Control},
  author={Zhao, Kaifeng and Li, Gen and Tang, Siyu},
  booktitle=ICLR,
  year={2024}
}

@article{chopin2023interaction,
  title={Interaction transformer for human reaction generation},
  author={Chopin, Baptiste and Tang, Hao and Otberdout, Naima and Daoudi, Mohamed and Sebe, Nicu},
  journal={IEEE Transactions on Multimedia},
  volume={25},
  pages={8842--8854},
  year={2023},
  publisher={IEEE}
}

@inproceedings{xu2024regennet,
  title={Regennet: Towards human action-reaction synthesis},
  author={Xu, Liang and Zhou, Yizhou and Yan, Yichao and Jin, Xin and Zhu, Wenhan and Rao, Fengyun and Yang, Xiaokang and Zeng, Wenjun},
  booktitle=CVPR,
  pages={1759--1769},
  year={2024}
}

@inproceedings{liu2024physreaction,
  title={PhysReaction: Physically plausible real-time humanoid reaction synthesis via forward dynamics guided 4d imitation},
  author={Liu, Yunze and Chen, Changxi and Ding, Chenjing and Yi, Li},
  booktitle={Proceedings of the 32nd ACM International Conference on Multimedia},
  pages={3771--3780},
  year={2024}
}

@inproceedings{ghosh2024remos,
  title={Remos: 3d motion-conditioned reaction synthesis for two-person interactions},
  author={Ghosh, Anindita and Dabral, Rishabh and Golyanik, Vladislav and Theobalt, Christian and Slusallek, Philipp},
  booktitle=ECCV,
  pages={418--437},
  year={2024},
  organization={Springer}
}

@inproceedings{
    tan2025think,
    title={Think Then React: Towards Unconstrained Action-to-Reaction Motion Generation},
    author={Wenhui Tan and Boyuan Li and Chuhao Jin and Wenbing Huang and Xiting Wang and Ruihua Song},
    booktitle=ICLR,
    year={2025}
}

@article{liang2024intergen,
  title={Intergen: Diffusion-based multi-human motion generation under complex interactions},
  author={Liang, Han and Zhang, Wenqian and Li, Wenxuan and Yu, Jingyi and Xu, Lan},
  journal=IJCV,
  volume={132},
  number={9},
  pages={3463--3483},
  year={2024},
  publisher={Springer}
}

@inproceedings{shafir2024human,
  title={Human Motion Diffusion as a Generative Prior},
  author={Shafir, Yoni and Tevet, Guy and Kapon, Roy and Bermano, Amit Haim},
  year={2024},
  booktitle=ICLR
}

@inproceedings{ruiz2024in2in,
  title={in2in: Leveraging individual information to generate human interactions},
  author={Ruiz-Ponce, Pablo and Barquero, German and Palmero, Cristina and Escalera, Sergio and Garc{\'\i}a-Rodr{\'\i}guez, Jos{\'e}},
  booktitle=CVPR,
  pages={1941--1951},
  year={2024}
}

@inproceedings{rot6d_zhou2019,
  title     = {{On the Continuity of Rotation Representations in Neural Networks}},
  author    = {Zhou, Yi and Barnes, Connelly and Lu, Jingwan and Yang, Jimei and Li, Hao},
  booktitle = CVPR,
  pages     = {5745--5753},
  year      = {2019}
}

@article{wang2024intercontrol,
  title={InterControl: Zero-shot Human Interaction Generation by Controlling Every Joint},
  author={Wang, Zhenzhi and Wang, Jingbo and Li, Yixuan and Lin, Dahua and Dai, Bo},
  journal=NIPS,
  volume={37},
  pages={105397--105424},
  year={2024}
}

@inproceedings{javedintermask,
  title={InterMask: 3D Human Interaction Generation via Collaborative Masked Modeling},
  author={Javed, Muhammad Gohar and Li, Xingyu and others},
  booktitle=ICLR,
  year={2024}
}

@inproceedings{xu2024inter,
  title={Inter-x: Towards versatile human-human interaction analysis},
  author={Xu, Liang and Lv, Xintao and Yan, Yichao and Jin, Xin and Wu, Shuwen and Xu, Congsheng and Liu, Yifan and Zhou, Yizhou and Rao, Fengyun and Sheng, Xingdong and others},
  booktitle=CVPR,
  pages={22260--22271},
  year={2024}
}

@inproceedings{cai2024digital,
  title={Digital life project: Autonomous 3d characters with social intelligence},
  author={Cai, Zhongang and Jiang, Jianping and Qing, Zhongfei and Guo, Xinying and Zhang, Mingyuan and Lin, Zhengyu and Mei, Haiyi and Wei, Chen and Wang, Ruisi and Yin, Wanqi and others},
  booktitle=CVPR,
  pages={582--592},
  year={2024}
}

@article{kingma2013auto,
  title={Auto-encoding variational bayes},
  author={Kingma, Diederik P and Welling, Max},
  journal={arXiv preprint arXiv:1312.6114},
  year={2013}
}

@article{yu2025socialgen,
  title={Socialgen: Modeling multi-human social interaction with language models},
  author={Yu, Heng and Zhang, Juze and Chen, Changan and Xiang, Tiange and Fang, Yusu and Niebles, Juan Carlos and Adeli, Ehsan},
  journal={arXiv preprint arXiv:2503.22906},
  year={2025}
}

@inproceedings{xu2025multi,
  title={Multi-Person Interaction Generation from Two-Person Motion Priors},
  author={Xu, Wenning and Fan, Shiyu and Henderson, Paul and Ho, Edmond SL},
  booktitle={Proceedings of the Special Interest Group on Computer Graphics and Interactive Techniques Conference Conference Papers},
  pages={1--11},
  year={2025}
}

@article{ota2025pino,
  title={PINO: Person-Interaction Noise Optimization for Long-Duration and Customizable Motion Generation of Arbitrary-Sized Groups},
  author={Ota, Sakuya and Yu, Qing and Fujiwara, Kent and Ikehata, Satoshi and Sato, Ikuro},
  journal={arXiv preprint arXiv:2507.19292},
  year={2025}
}
